\pdfoutput=1 
\documentclass[english,ngerman]{template/alr_report}

\renewcommand{\mythesis}{\mastersthesis}
\renewcommand{\mytitle}{Optimizing the Long-Term Behaviour of Deep Reinforcement Learning for Pushing and Grasping}
\renewcommand{\myshorttitle}{} 
\renewcommand{\myname}{Rodrigo Chau}
\renewcommand{\timestart}{January 19\textsuperscript{th}, 2021}
\renewcommand{\timeend}{August 19\textsuperscript{th}, 2021}

\newcommand{\referee}{Prof. Dr. Techn. Gerhard Neumann}
\newcommand{\refereetwo}{Prof. Dr. Ing. Tamim Asfour}
\newcommand{\advisor}{Dr. Hanna Ziesche }
\newcommand{\advisortwo}{Dr. Vien Ngo}

\hypersetup{
	pdfauthor={\myname},
	pdftitle={\mytitle},
	pdfsubject={\mytitle},
	pdfkeywords={Reinforcement Learning, Robotics, Deep Learning, Q-Learning}  
}

\DeclareMathOperator*{\argmax}{argmax}

\begin{document}

\selectlanguage{english}

\frontmatter

\maketitle
\cleardoublepage


\chapter*{Zusammenfassung}
\addcontentsline{toc}{chapter}{Zusammenfassung}

In den letzten Jahren gab es mehrere Veröffentlichungen im Gebiet des Robot Reinforcement Learning, in denen Forscher neuronale Netze, die für Computer Vision Aufgaben entwickelt wurden, in Reinforcement Learning Algorithmen einsetzen. Üblicherweise werden diese Netze verwendet um die Beobachtungen der Umgebung des Roboters zu verarbeiten. Wir untersuchen zwei Beispiele für diesen Ansatz, das \glqq Visual Pushing for Grasping\grqq{} (VPG) System von \citeauthor{Zeng2018} und das \glqq Hourglass\grqq{} System von \citeauthor{Ewerton2021}, welches eine Weiterentwicklung des Ersteren ist. Diese Systeme sind Instanzen von Deep Q-Learning-Algorithmen und trainieren daher ihre Netze die Q-Werte eines Problems möglichst genau vorherzusagen. Das Hauptmerkmal ihrer Systeme ist die enge Beziehung zwischen den Bildern der Umgebung und dem Aktionsraum des Roboters, jedes Pixel repräsentiert eine mögliche Aktion. Im Fokus unserer Arbeit steht die Untersuchung der Fähigkeit dieser Ansätze langfristige Belohnungen und Strategien vorherzusagen. Das von \citeauthor{Zeng2018} untersuchte Problem erfordert nur ein begrenztes Maß an Vorraussicht des Agenten. \citeauthor{Ewerton2021} erreichen die beste Leistung mit einem Agenten, der nur die unmittelbarste Aktion in Betracht zieht. Für unsere Untersuchung entwickeln wir ein neues Sortierproblem und Belohnungsfunktion. Unser Problem erfordert, dass Agenten zukünftige Belohnungen genau schätzen und hohe Diskontfaktoren in der Berechnung des Q-Werts verwenden. Wir testen das Verhalten einer Adaption des VPG Systems mit unserer Aufgabe. Wir zeigen, dass diese Adaption die erforderlichen langfristigen Aktionssequenzen nicht genau vorhersagen kann. Neben den von \citeauthor{Ewerton2021} aufgezeigten Einschränkungen zeigt es das im Deep Q-Learning bekannte Problem von überschätzten Q-Werten. Um unsere Aufgabe zu lösen, wenden wir uns den Hourglass Modellen zu und kombinieren sie mit dem Double Q-Learning Ansatz. Dies ermöglicht den Modellen langfristige Aktionssequenzen zu prädizieren, wenn sie mit großen Diskontfaktoren trainiert werden. Unsere Ergebnisse zeigen, dass Double Q-Learning für das Training mit sehr hohen Diskontfaktoren unerlässlich ist, da die Q-Werte sonst divergieren. Wir experimentieren außerdem mit verschiedenen Ansätzen für Diskontierung während des Trainings, Verlustfunktion und Explorationsverfahren. Unsere Ergebnisse zeigen, dass diese die Leistung des Modells für unsere Aufgabe nicht sichtbar beeinflussen. 

\cleardoublepage

\iflanguage{english}{%

\chapter*{Abstract}
\addcontentsline{toc}{chapter}{Abstract}

In recent years there have been multiple publications in the field of Robot Reinforcement Learning, where researchers employ neural networks, which were previously used in Computer Vision tasks, in Reinforcement Learning algorithms. Commonly, researchers use these networks to process the observations of the robot's environment. In our work, we examine two examples of this approach. We investigate the "Visual Pushing for Grasping" (VPG) system by \citeauthor{Zeng2018} and the "Hourglass" system by \citeauthor{Ewerton2021}, an evolution of the former. These systems are instances of Deep Q-Learning algorithms and therefore aim to train their networks to accurately predict the Q-Values of a Reinforcement Learning problem. The key characteristic of their systems is the close relation between the images of the workspace and the action space of the robot, each pixel of the image represents a possible action. The focus of our work is the investigation of the capabilities of both systems to learn long-term rewards and policies. \citeauthor{Zeng2018} original task only needs a limited amount of foresight. \citeauthor{Ewerton2021} attain their best performance using an agent which only takes the most immediate action under consideration. We are interested in the ability of their models and training algorithms to accurately predict long-term Q-Values. To evaluate this ability, we design a new bin sorting task and reward function. Our task requires agents to accurately estimate future rewards and therefore use high discount factors in their Q-Value calculation. We investigate the behaviour of an adaptation of the VPG training algorithm on our task. We show that this adaptation can not accurately predict the required long-term action sequences. In addition to the limitations identified by \citeauthor{Ewerton2021}, it suffers from the known Deep Q-Learning problem of overestimated Q-Values. In an effort to solve our task, we turn to the Hourglass models and combine them with the Double Q-Learning approach. We show that this approach enables the models to accurately predict long-term action sequences when trained with large discount factors. Our results show that the Double Q-Learning technique is essential for training with very high discount factors, as the models Q-Value predictions diverge otherwise. We also experiment with different approaches for discount factor scheduling, loss calculation and exploration procedures. Our results show that the latter factors do not visibly influence the model's performance for our task.

  \cleardoublepage%
}{}

\tableofcontents
\cleardoublepage

\mainmatter


\chapter{Introduction}
Artificial intelligence (AI) is one of the most vibrant fields of research in today's time. From self-driving cars, to autonomous factories and robotic butlers, the possibilities that arise from complex-reasoning machines seem endless. Fueled by advances in the processing speed of large amounts of data, researchers have achieved many milestones in AI in the last decade. A couple of those achievements have garnered the attention of mainstream media due to the fact that they have surpassed human-level performance in their respective task. Two of the most popular examples are the deep neural network "ResNet", which was able to attain human-level scores on the ImageNet classification task in 2015 \citep{ResNet} and the "AlphaGo" program which was able to beat the number one ranked Go player in 2017 \citep{AlphaGo}. 

The ImageNet classification task is an established benchmark task for image classification systems in the field of Computer Vision \citep{ImageNet}. In Computer Vision researchers develop computer systems which can form an understanding of digital images. These systems receive images as input and output the information of interest to their user. In image classification the goal is to localize and label the contents of an image. Historically, systems used a combination of hand-crafted filters and pattern recognition algorithms to solve this task. With breakthroughs in the field of machine learning, especially the fast training of deep neural networks, researchers were able to design systems which exceeded the then state-of-the-art performance by a large margin. Rather than relying on hand-crafted filters, these systems learn to identify the relevant features of an image. For image classification, one trains neural networks with supervised learning algorithms. Essentially, these algorithms repeatedly compare the network output for a given image to the underlying ground truth and modify the parameters of the network based on the difference, or error, between the two. In reference to the large number of layers of the neural networks in use, this field of research is also called Deep Learning \citep{Goodfellow2016}.

The development of the AlphaGo program \citep{AlphaGo} falls into the category of Reinforcement Learning \citep{Sutton2018}. In Reinforcement Learning the goal is to train an agent to execute a sequence of good actions. The agent learns about the notion of good or bad actions through trial and error. It receives a reward if the action was good and possibly a penalty if it wasn't. As a general machine learning algorithm, one can apply Reinforcement Learning to many different problems. One models the agent's environment as a Markov Decision Process, consisting of states and actions. The agent takes actions in this environment, which then transitions from one state to another. The most interesting tasks require strategies which consist of multiple consecutive actions. For these problems the agent has to learn to plan long-term strategies by drawing the correct conclusions from the observations of the current state of the environment. This becomes more and more difficult, as the number of states and actions increases. For the Go board game the number of legal positions of a regular sized board exceeds $2\times10^{170}$ \citep{Tromp2007}. The AlphaGo developers approached this problem by training a deep neural network to estimate the winning percentage of different moves and another to select moves. As such it is a prime example for Deep Reinforcement Learning, the combination of Deep Learning and Reinforcement Learning.

Our field of study is Deep Robot Reinforcement Learning, where the agent in question is a robot which has to solve a task by interacting with its environment. Commonly, these tasks consist of manipulating objects of interest in the environment. Recently, there have been several publications which approach Robot Reinforcement Learning problems with deep learning methods derived from Computer Vision solutions. A recurring approach that one can notice in these publications is the usage of deep neural network architectures, which have been tried and tested in an Computer Vision setting, to process the observations of the robots environment. One example of this approach is the "Grasp2Vec" system which uses ResNet to learn object-centric representations of its environment \citep{Grasp2Vec}. Another example is the "Visual Pushing for Grasping" (VPG) system \citep{Zeng2018}. It builds on the capabilities of another image classification neural network, DenseNet \citep{DenseNet}, to solve a bin picking task. A novelty of the VPG approach is its strategy to reduce the amount of actions the system has to consider, by reducing them to a set of pre-defined motions which are represented by images. \citeauthor{Zeng2018}'s main focus is showing that their system can learn two different action types, pushing and grasping, jointly and leverage synergies between the two. They demonstrate the suitability of their approach for a simple task which requires only a limited amount of planning to be solved.

\citet{Ewerton2021} built on the VPG system in the development of their own robot pushing system. They show that replacing DenseNet with an Hourglass network architecture, which \citet{Newell2016} originally introduced for human pose estimation, proves to be beneficial. In their task the robot has to push the objects from a table into a box. While this is a good example for a task which requires some amount of planning, their most successful configuration uses a strategy which only takes the single, most immediate action into consideration. The authors state the informative reward function as a possible reason for this behaviour.

The VPG and Hourglass systems are instances of Deep Q-Learning algorithms. In Deep Q-Learning one approaches the reinforcement learning problem by training a deep neural network to accurately predict the Q-Values \citep{Mnih2015}. A Q-Value is the sum of the reward for the most immediate action and an estimate of the future reward that the agent will subsequently receive. To prevent infinite Q-Values, one multiplies the future reward with a discount factor. This factor determines how far the agent looks into the future when making its decisions. For low discount factors it is more focused on maximizing the immediate reward. For high discount factors it will accept lower immediate rewards and even penalties, if they lead to higher rewards in the future. When the system is able to predict the true Q-Value to a sufficient degree, it can generate the optimal strategy by always selecting the action with the maximum Q-Value for a given state. A known problem for Deep Q-Learning algorithms is the overestimation of Q-Values \citep{VanHasselt2016}. This overestimation can lead to suboptimal agent decisions. A method to mitigate this problem is the introduction of the Double Q-Learning approach \citep{VanHasselt2016}.
 
In this master thesis, we investigate the suitability of the VPG and Hourglass approaches for a different task which requires long-term strategies to solve. We are interested in their capability to apply long-term reasoning, even when the agents only receive rewards occasionally. We show that one requires large discount factors to solve these kinds of problems with Deep Q-Learning. Furthermore, we show that a requirement for using large discount factors is dealing with the problem of overestimated Q-Values, by using Double Q-Learning.

Our practical work consists of implementing the new task in a MuJoCo simulation environment \citep{mujoco}. We build on our previous work in the practical Autonomous Learning Robots course, where we realized \citeauthor{Zeng2018}'s original task in MuJoCo \citep{Chau2020}. We implement the new task using the OpenAi gym interface \citep{openaigym}. We also adapt both the VPG and Hourglass approach to the Stable Baselines3 interface \citep{stable-baselines3}. The Stable Baselines3 package offers a reinforcement learning algorithm framework, including a replay buffer and target network implementation. We extend it with the Double Q-Learning method and prioritized experience replay buffers \citep{Schaul2016}. We train multiple configurations of our adaptations, trying out different parameters. We test the trained systems and analyze the results.

In our written report, we start by introducing the fundamentals of Deep Reinforcement Learning. We then describe the VPG and Hourglass system in detail. We outline the design of our task, including its reward function. We describe the adaptations we made to the VPG system and the training procedure. We present the test results of the trained systems, analyze and discuss them. In particular, we look at its overestimated Q-Values and bias towards certain actions. We then describe the modifications we made to the Deep Q-Learning system to optimize its behaviour on our task. These modifications include the replacement of the VPG network with the Hourglass network, the usage of a target network and the Double Q-learning technique. We describe our changes to the exploration and training procedure. We present and evaluate the test results of the optimized systems. Our main focus is comparing different discount factors and systems with and without target networks and Double Q-Learning. We also analyze the impact of discount factor scheduling \citep{Francois-Lavet2015}, different exploration techniques and loss functions on the performance of the systems. Lastly, we draw conclusions and outline possible future work.
\cleardoublepage


\chapter{Fundamentals}
In this chapter we describe the theoretical fundamentals of our work. We first outline the foundations of Reinforcement Learning \citep{Sutton2018} and Q-Learning \citep{watkins1992}, introduce neural networks \citep{Goodfellow2016} and their application in Q-Learning \citep{Lin1992}. We then describe three techniques relevant for Deep Q-Learning: Experience Replay \citep{Lin1992}, Target Networks \citep{Mnih2015} and Double Q-Learning \citep{VanHasselt2016}. Lastly, we briefly describe MuJoCo \citep{mujoco}, the simulation engine we used in our experiments.

\section{Reinforcement Learning}
In this section we follow the definitions from \citet{Sutton2018} and refer to the same for a more detailed introduction. Reinforcement Learning is an area of machine learning which deals with training an agent to perform a rewarding sequence of actions in an environment. It distinguishes itself from supervised or unsupervised machine learning as a third basic machine learning paradigm.

An agent can be anything from a virtual or simulated entity to a robot in the real world. When an agent faces a problem that requires some form of inference from observations of an environment and planning a strategy of actions, one can train it with reinforcement learning. The key aspect of reinforcement learning is to give the agent a reward for the executed actions and training it to maximize the cumulative reward. 

Formally, one models environments for reinforcement learning problems as Markov Decision Processes (MDPs). Note that the following definitions assume discrete-time environments and MDPs. An MDP consists of a set of States $S$, a set of actions $A$, a reward function $R(s_t,a_t)$, a transition model $P(s_{t+1}|s_t,a_t)$ and an initial state distribution $\mu_0(s)$. The key, name giving, "Markov Property" of an MDP is the so-called memorylessness, i.e., the independence of the transition probability from all past states and actions given the last state-action pair. This property is defined by
\begin{equation*}\label{eq:1}
    p(s_{t+1}|s_t,a_t,s_{t-1},a_{t-1},s_{t-2},a_{t-2},...) = p(s_{t+1}|s_t,a_t).
\end{equation*}
An agent interacts with the environment in a given state $s_t$ according to a policy $\pi(a|s)$, which defines which action to take in that state. A non-deterministic policy can also define probabilities for different actions for a given state. The environment then transitions to a new state $s_{t+1}$ according to the transition model $P$. The agent receives a reward $r_t$ as defined by the reward function. The goal of reinforcement learning is to find an optimal policy $\pi^*$ that maximizes the expected discounted reward starting from timestep $t$, which is
\begin{equation*}\label{eq:2}
    G_{t} := \sum_{k=0}^\infty \gamma^{\,k } r_{t+k}.
\end{equation*}
The discount factor $\gamma \in [0;1)$ represents the trade-off between optimizing short-term and long-term reward. One uses high $\gamma$ values to optimize the long-term reward and low $\gamma$ values to optimize the short term reward.

Two important functions which link states and state-action-pairs to the expected reward are the V-Function and the Q-Function. The V-Function
\begin{equation*}\label{eq:3}
    V: S \to \mathbb{R}, \quad V_{\pi}(s) := \mathbb{E}_\pi [\sum_{k=0}^\infty \gamma^{\,k} r_{t+k} | s_t = s]
\end{equation*}
represents the expected future reward, if an agent is in state $s$ and follows the policy $\pi$ in subsequent steps. As such it serves as a quality metric for states, i.e., the V-Value is high, if the agent can gather high rewards starting from state $s$.

The Q-Function
\begin{equation*}\label{eq:4}
    Q: S \times A \to \mathbb{R}, \quad Q_{\pi}(s,a) := \mathbb{E}_\pi [\sum_{k=0}^\infty \gamma^{\,k} r_{t+k} | s_t = s, a_t=a]
\end{equation*}
similarly represents the expected reward, if an agent is in state $s$, executes action $a$ and follows the policy $\pi$ afterwards. Therefore, the Q-Function is a quality metric for one explicit state-action pair and measures how promising it is to take a specific action in a certain state.

There are various different reinforcement learning algorithms. \citet{Dong2020} distinguish them by a couple of key characteristics. First, one can differentiate between model-based and model-free approaches. Model-based approaches either utilize a given MDP model or try to learn a model of the problem at hand. Model-free approaches do not need such a model but aim to search for the highest reward by interacting with the environment. They are further categorized into value-based and policy-based methods. Policy-based methods operate directly on the policy and seek to improve it iteratively in order to maximize the gathered reward. Value-based methods aim to optimize an estimate of the Q-Function to approach its true value. Lastly, value-based methods are divided into on-policy and off-policy methods. On-policy methods optimize the same policy that is used to make decisions in an environment. Off-policy methods train a different, separate policy than the one used to generate experiences.

\section{Q-Learning}
One model-free, value-based, off-policy algorithm is Q-Learning \citep{watkins1992}. As the name suggests, it is centered around the aforementioned Q-Function. If one knows the Q-Function, one can use it to generate the optimal policy by always choosing the action with the highest associated Q-Value, i.e.,
\begin{equation*}\label{eq:5}
    \pi^*(s)= \argmax_a \, Q(s,a).
\end{equation*}
As one rarely knows the Q-Function beforehand, the challenge becomes finding a good Q-Function estimate. Q-Learning iteratively refines a Q-Function estimate with experience gathered from interacting with the environment. The algorithm starts by setting the Q-Function $Q(s,a)$ for all state-action-pairs to some arbitrary or pre-chosen value. The algorithm then initializes an episode with an initial state $s_t = s_0$ of the problem. It chooses an action $a_t$ and executes it. The environment then transitions to the following state $s_{t+1}$. For this action the agent receives a reward $r_t$. The algorithm updates the Q-Value estimate with the function 
\begin{equation*}
    Q(s_t,a_t) = Q(s_t,a_t) + \alpha [r_{t} + \gamma \, \max_a Q(s_{t+1},a) - Q(s_t,a_t)],
\end{equation*}
where $\alpha \in (0,1]$ is the learning rate. The algorithm repeats this process in a loop for the subsequent states until the episode terminates, after which it starts a new episode.

One main challenge of reinforcement learning algorithms, including Q-Learning, is balancing exploitation and exploration when choosing the action $a_t$. Exploitation aims to leverage the gained knowledge to maximize the received reward. In the context of Q-Learning this means greedily choosing the action with the highest associated Q-Value when generating new samples. On the other hand, exploration seeks to gather new knowledge by selecting other actions, potentially leading to higher rewards. One has to balance these two strategies for Q-Learning to succeed. In practice, Q-Learning algorithms often use an "$\epsilon$-greedy" strategy. This strategy selects a random action with a probability of $\epsilon \in (0;1)$ and a greedy action otherwise.

The Q-Learning algorithm runs until it has met some termination condition, e.g., when it reaches pre-defined number of $n$ update steps. Ideally, the final Q-Function estimate is a good enough approximation and can be used to extract the optimal policy. \citet{watkins1992} prove that the Q-Function converges for $n \to \infty$ if the following conditions are met: 1) Every state-action pair is visited infinitely often. 2) Given bounded rewards, the sum of the learning rates diverges, while the sum of their squares converges. The first condition is satisfied if each action has a non-zero selection probability in any given state, as is the case for $\epsilon$-greedy exploration strategies. The second condition signifies that the learning rate has to approach a small enough value over training, but not too quickly.

One of the main benefits of Q-Learning is that it is model-free, i.e., one does not need a model of the underlying MDP to apply Q-Learning. In its standard form Q-Learning is only applicable in tabular problems, i.e., those with bounded, discrete state and action spaces. Meeting the convergence criteria is increasingly difficult, as the sizes of state and action spaces increase and is impossible if they are unbound or continuous. In these cases, it is possible to combine Q-Learning with function approximation, e.g., by using artificial neural networks \citep{VanHasselt2012}.

\section{Neural Networks}
Artificial neural networks are the primary catalyst of recent advances in the field of machine learning, including reinforcement learning. In this section we follow the definitions from \citet{Goodfellow2016}. At their lowest level aritificial neural networks are composed of artificial neurons. An artificial neuron can receive input from any number of incoming connections. The neuron combines these inputs $x_n$ as 
\begin{equation*}
    y_k = \varphi \left(\sum_{j=0}^{m} w_{j}x_j\right) \textrm{, with $x_0 = 1$ const.},
\end{equation*}
where $w_j$ is the weight of the corresponding input. The weight $w_0$ is also called the bias $b$ and $\varphi$ is an activation function which generates the output $y_k$. The neuron then propagates this output to all outgoing connections.

One of the basic neural network architectures is the multilayer perceptron (MLP). MLPs consist of artificial neurons which are grouped into layers. The first layer receives the input information, e.g., digitized images, voice recordings or documents. From here the information passes through possibly multiple hidden layers. The last layer produces the output of the network. The network graph is acyclic, i.e., connections are only allowed in one direction, so that there exist no cycles in the network graph. Furthermore, neurons in the same layer are not connected, but only receive information from previous layers and send their output to subsequent layers. As such MLPs form a class of "feedforward" artificial neural networks. If every neurons of a layer are connected to every neurons of the subsequent layer, the network is called "fully connected". Networks which have a large number of layers are also referred to as "deep networks".

As universal function approximators, MLPs can be used for a wide variety of tasks. They are most known for their application to classification tasks, e.g., image classification. For this task the network receives an input image, and outputs a prediction for the corresponding class, e.g., 'cat' or 'dog'. In this setting, one trains MLPs using supervised learning with a large batch of labeled training data. In reference to the deep network architectures this process is also called deep learning. The training algorithm repeatedly compares the output of the network with the desired label for the training data. Based on the difference of the output and the label, it calculates an error. With the derivative of each neuron's weights with respect to the error, the algorithm backpropagates the error through the network, updating the weights. After it performs enough backpropagation, the network function ideally converges to the desired target. One can evaluate the performance of the trained network by evaluating it on an unseen data set.

For image classification one usually employs a class of MLPs called Convolutional Neural Networks (CNNs). CNNs have layers which apply the convolution operation on their input. In the discrete case, the convolution operation is essentially a weighted sum of the input. For the 2-D, single channel inputs the convolutional layers define a kernel consisting of $n\times m$ weights. For a single output element the layer combines the input elements in the corresponding area by computing the weighted sum and adding a bias value. For the next output element, the kernel is then moved along the input dimension by the stride, a preset amount of elements. One can pad the input image before the convolutional layer processes it. For multiple input channels, the kernel has the size of $C_{\textrm{in}} \times n \times m $ as combines the input elements across all input channels. In the case of multiple output channels an individual kernel is used for every output channel. The training algorithm adjusts the weights of the convolutional layers through backpropagation. The advantage of convolutional layers is that it reduces the number of parameters needed when compared to a fully connected layer. They are also translation equivariant, as they reuse the same kernel across the entire input. This is especially important in image classification, as it should not matter where a distinguishing feature is located in an image to be able to recognize it.

One needs many different techniques and fine-tuning steps to successfully train a neural network. Covering all of these would exceed the scope of this work. Again, we refer to \citet{Goodfellow2016} for further reading.

\section{Q-Learning with Neural Networks} \label{sec:deepQ}
As mentioned, standard Q-Learning \citep{watkins1992} is only applicable in tabular cases, i.e., when the state and action spaces are bound and discrete. In other cases, one needs to deal with the problem that continuous and unbound spaces pose \citep{VanHasselt2012}. One would be inclined to solve this problem by using neural networks as function approximators for the Q-Function. In this section, we describe a naive version of standard Q-Learning with neural networks, similar to the "QCON" framework from \citet{Lin1992}.

The neural network Q-Function approximation is denoted as $Q_\theta(s,a)$, where $\theta$ represent the network weights. This network is also called Q-Network. As with standard Q-Learning, the goal is to find a good enough estimate of the true Q-Function. The Q-Network receives the inputs $s_t$ and $a_t$ and outputs its current estimate $Q_\theta(s_t,a_t)$. For a given experience $e_t=(s_t,a_t,r_t,s_{t+1})$ the algorithm compares this estimate to a target value $y_t$, given by
\begin{equation*}
    y_t = r_t + \gamma \, \max_{a_{t+1}} Q_\theta(s_{t+1}, a_{t+1}).
\end{equation*}
This value is the sum of the immediate reward $r_t$ and the estimate of the future reward, discounted with $\gamma$. Note that the future reward estimate is given by the maximum value of $Q_\theta$, the Q-Networks estimate for the subsequent state. The algorithm then computes a loss by applying some loss function, e.g. the mean squared error (MSE) loss
\begin{equation*}
    \mathcal{L}_t = (Q_\theta(s_t,a_t)-y_t)^2.
\end{equation*}
The algorithm then updates the Q-Function estimate by backpropagating this loss through the Q-Network, updating the weights with gradient descent, i.e., applying
\begin{equation*}
    \theta' = \theta - \alpha \frac{d\mathcal{L}_t}{d\theta},
\end{equation*}
with the learning rate $\alpha$.
The most basic Q-Learning algorithm generates one experience in each timestep and adjusts the approximation based on this experience. It repeats this method for the subsequent states, until some termination condition is met. 

\citet{Tsitsiklis1997} show that temporal difference methods, e.g., Q-Learning, diverge when using non-linear function approximators, e.g., neural networks. The main problems are the correlation between the sequential states and the fact that the target value $y_t$ changes as the Q-Function estimate $Q_\theta$ changes. There are two central strategies to tackle these problems, Experience Replay \citep{Lin1992} and Target Q-Networks \citep{Mnih2015}.

\section{Experience Replay} \label{sec:experience_replay}
Experience Replay uses a replay buffer to store the agent's experiences \citep{Lin1992}. The buffer has a capacity of $N$. During learning, instead of only using the most recent experience in each time step, the Q-Network trains on a mini-batch of $n$ samples from the replay buffer. One advantage of this approach is that it decreases the correlation of sampled values, as it can sample experiences which were not made in close succession. Another benefit is the increased data-efficiency of the algorithm, as it is able to reuse experiences multiple times. In its basic form Experience Replay selects sample from the buffer uniformly at random. This selection strategy can be limiting, if rewards are rare and the bulk of the buffer therefore contains unsuccessful experiences. \citet{Schaul2016} show that it is beneficial to employ prioritization in this case. They use the temporal difference error
\begin{equation*}
    \delta_t := r_t + \gamma \, \max_{a_{t+1}} Q_\theta(s_{t+1}, a_{t+1}) - Q_\theta(s_{t}, a_{t})
\end{equation*}
as a prioritization metric and a prioritized stochastic sampling method based on this value. The probability of sampling a specific transition $i$ is given by
\begin{equation*}
    P(i) = \frac{p_i^\alpha}{\sum_k p_k^\alpha}.
\end{equation*}
$p_i$ is either proportional to $\delta$ or rank-based and $\alpha$ is a parameter which defines the extent of prioritization. In rank-based prioritization
\begin{equation*}
p_i = \frac{1}{\textrm{rank}(i)},
\end{equation*}
where $\textrm{rank}(i)$ is the position of transition $i$ in the replay buffer sorted by $|\delta_i|$. In this case, the resulting probability distribution $P$ is equivalent to a power-law distribution with the exponent $\alpha$.

\section{Target Network}
The naive application of the Q-Learning algorithm on Q-Networks trains them on the target values
\begin{equation*}
    y_t = r_t + \gamma \, \max_{a_{t+1}} Q_\theta(s_{t+1}, a_{t+1}).
\end{equation*}
As \citet{Tsitsiklis1997} show, the fact that updating $Q_\theta$ possibly changes the target $y_t$ can lead to instability and divergence of the Q-Function estimate. \citet{Mnih2015} published a variant of the Q-Learning algorithm, called Deep Q-Network (DQN), which aims to deal with this issue. They introduce a second neural network called target network. This network has the same architecture as the Q-Network. $\theta^-$ denotes the weights of this target network and $Q_{\theta^-}$ denotes its Q-Function approximation. They initialize the target network with the same random initial weights as the Q-Network i.e., $\theta^- = \theta$. During training, the algorithm calculates the future reward estimation in the target label based on the prediction of the target network, i.e.,
\begin{equation*}
    y_t = r_t + \gamma \, \max_{a_{t+1}} Q_{\theta^-}(s_{t+1}, a_{t+1}).
\end{equation*}
The algorithm keeps the target network unchanged for a pre-defined number of steps $C$. Every $C$ steps it updates the target network assigning $\theta^- = \theta$. This approach ensures that the effect of changing $Q_\theta$ on the target label values is delayed. This reduces the likelihood of divergence and oscillations during training.

\section{Double Q-Learning}\label{sec:doubleQ}
While target networks help to reduce the impact of changing the Q-Function estimate on the target label, \citet{VanHasselt2016} show that the DQN algorithm is prone to overestimate the values of the true Q-Function. The max-operator in the target label formula is likely to select overestimated values. The authors solve this problem by combining the Double Q-Learning approach \citep{vanHasselt2010} with DQN. Double Q-Learning decouples action selection and evaluation by using a different set of weights for each. The resulting algorithm, "Double DQN", uses the primary Q-Network weights $\theta$ for action selection and the target network weights ${\theta^-}$ to evaluate the action. Therefore, it calculates the target label as
\begin{equation*}
    y_t = r_t + \gamma \, Q_{\theta^-}(s_{t+1}, \, \argmax_a Q_\theta(s_{t+1}, a)).
\end{equation*}
The authors show in their results that the algorithm exhibits more stability and reliability than the original DQN algorithm.

\section{MuJoCo Simulation}
MuJoCo is a physics engine developed by Emo Todorov for \citet{mujoco}. MuJoCo is an acronym for "Multi-Joint dynamics with Contact". According to the authors the engine was built to satisfy demanding speed, stability and accuracy requirements for research in the field of optimal control. MuJoCo generates a scene based on specifications in XML. It supports both off- and on-screen rendering. It is originally written in C, but has been made available for use with Python by the mujoco-py package of the \citet{mujoco-py}. The ALR institute at the KIT has developed a simulation framework which we use. It offers pre-defined scenes including a Franka Panda robot. It has an automatic XML generation mechanism to define the scenes from Python objects. It also supports virtual cameras with image and point cloud capturing.

\cleardoublepage


\chapter{Related Work}

In this chapter we first describe the "Visual Pushing for Grasping" (VPG) system \citep{Zeng2018}. The VPG system is an example of a DQN algorithm applied to a robotic manipulation task. We then outline the Hourglass system \citep{Ewerton2021}. This system is an evolution of the VPG approach with modifications to key components, which the authors made to remove different limitations of the original system. 

\section{Visual Pushing for Grasping System}
\citet{Zeng2018} introduced the VPG system in the paper "Learning Synergies between Pushing and Grasping with Self-Supervised Deep Reinforcement Learning". In their task, a robotic arm has to pick up individual objects from a cluttered heap of objects. The robot is able to perform two kinds of actions, grasping and pushing. Grasping objects successfully is the primary goal. Pushing objects should enable the robot to subsequently grasp with a higher success rate. Their method is a DQN variant which centers around training two networks, one for each action type, at the same time. They demonstrate that this joint training technique enables the robot to learn synergies between the two action types.

\citeauthor{Zeng2018} model their task as a discrete-time MDP. They use RGB-D heightmaps to represent the states of their environment. In each timestep an RGB-D camera takes a $224 \times 224$ image of the workspace. The system transforms these images, so that the depth value of each pixel $p$ is the corresponding point's height from the workspace surface. Their actions consist of two parameters, the motion primitive $\psi$ and the 3D location $q$, i.e.,
\begin{equation*}
    a=(\psi, q) \, | \, \psi \in \{\textrm{push,grasp}\}, q \twoheadrightarrow p \in s_t.
\end{equation*}
A key characteristic of their approach is mapping the pixels $p$ of the state representation $s_t$ to 3D action locations $q$. Each pixel represents a different location in the workspace. The world coordinates of this location can be calculated by,
\begin{align*}
    x_{\textrm{world}} &= X_{\min} + p_x \frac{X_{\max} - X_{\min}}{R_x}\\
    y_{\textrm{world}} &= Y_{\min} + p_y \frac{Y_{\max} - Y_{\min}}{R_y}\\
    z_{\textrm{world}} &= Z_{\min} + d_p,
\end{align*}
where $(X,Y,Z)_{\min}$ and $(X,Y,Z)_{\max}$ are the workspace minimum and maximum boundaries in world coordinates for the respective axis. $R_x$ and $R_y$ are the $x$- and $y$-resolution of the heightmap, $p_x$ and $p_y$ are the pixels indices in the image and $d_p$ is its associated depth value. Additionally, the authors define $k = 16$ rotations for their grasping and pushing actions. They define their push motion primitive with the starting point $q$ and the direction of the push. The length of a push is always $10$cm along a straight trajectory. For grasps, $q$ represents the center point of a top-down grasp. Here, the rotation represents the orientation of the parallel-jaw gripper.

This approach is a form of action space discretization, where, depending on the current environment state, a different set of actions can be executed. It also limits the amount of actions that have to be evaluated by the Q-Network. In total, the action space consists of $k \times R_x \times R_y = 1,605,632$ actions at each timestep.

The authors define two reward functions, one for each of their systems action primitives. The reward function for grasping actions is
\begin{equation*}
    R_g(s_t, a_t, s_{t+1}) = 
    \begin{cases}
    1, & \textrm{if $a_t$ is a grasp and was successful,}\\
    0, & \textrm{otherwise}.
    \end{cases}
\end{equation*}
They check for a successful grasp, by measuring the distance between the gripper fingers after a grasp attempt and comparing it to a pre-defined threshold.
For pushing actions, the reward function is
\begin{equation*}
    R_p(s_t, a_t, s_{t+1}) = 
    \begin{cases}
    0.5, & \textrm{if $a_t$ is a push and lead to a detectable change,}\\
    0, & \textrm{otherwise}.
    \end{cases}
\end{equation*}
The system detects a change if $\sum(s_{t+1}-s_t) > \tau$, i.e., if the sum of differences between the depth channels of the heightmaps representing $s$ and $s_{t+1}$ is greater than a pre-defined threshold $\tau$.

\citeauthor{Zeng2018}'s system receives the state representation as input and outputs a Q-Value map for each action and orientation. Tow individual networks, one for each action type, evaluate the respective actions. Each network receives 16 rotated heightmaps, where each rotation represents one individual action orientation. Equivalently, each network outputs 16 Q-Value maps, i.e., $224 \times 224$ matrices of Q-Values where the matrix indices correspond to the x and y indices of the input images. The entry with the highest value across all 32 Q-Value maps represents the action with the highest estimated Q-Value, i.e., the action which the networks predicts to be the most rewarding.

The network architecture is the same for both networks: 
\begin{quote}
Two parallel 121-layer DenseNet pre-trained on ImageNet, followed by channel-wise concatenation and 2 additional $1\times1$ convolutional layers interleaved with nonlinear activation functions (ReLU) and spatial batch normalization, then bilinearly upsampled. One DenseNet tower takes as input the color channels (RGB) of the heightmap, while the other takes as input the channelwise cloned depth channel (DDD) (normalized by subtracting mean and dividing standard deviation) of the heightmap. (\citet{Zeng2018}, p.4) 
\end{quote} 
The authors zero-pad and zoom their images before rotating them. This pre-processing procedure results in an input resolution of $640\times640$. Before the system upsamples the Q-Networks output, it reverts the image rotation. At this stage, the Q-Value map resolution is $20x20$ and the upsampling factor is $16$. After upsampling, the extra padding is removed, leading to the desired output resolution of $224\times224$.

After initialization, the training algorithm selects an action for the current state $s_{i}$ of the environment using an $\epsilon$-greedy strategy. For exploitation it selects the action with the highest associated Q-Value. For exploration, it chooses between pushing and grasping at random, with equal probability. Noticeably, it also selects the action with the highest Q-Value from the chosen action type. The algorithm executes the selected action $a_i$ in a parallel thread and evaluates its success. It computes the reward $r_i$, as described above, and with it the target label value
\begin{equation*}
    y_i =
    \begin{cases}
    r_i + \gamma \, \max\limits_{a} Q_{\theta_i}(s_{i+1},a), & \textrm{if the action was successful or lead to a change,} \\
    r_i, & \textrm{otherwise}.
    \end{cases}
\end{equation*}
Notably, the target label calculation includes a future reward in the case of a successful push or grasp, or if a grasp was unsuccessful but lead to a detectable change. In other cases, the target label does not include a future reward term, but is actually always $y_i = r_i = 0$. Their code published in \citep{VPGGithub} does not use a target network for the future reward calculation. The algorithm saves the current experience $e_i = (s_i, a_i, s_{i+1}, Q_{\theta_i}(s_i,a_i),r_i, y_i)$ to the replay buffer. It then calculates the loss using the Huber loss function
\begin{equation*}
    \mathcal{L}_i =
    \begin{cases}
    \frac{1}{2}(Q_{\theta_i}(s_i,a_i) -  y_i)^2, & \textrm{for } |Q_{\theta_i}(s_i,a_i) -  y_i| < 1, \\
    |Q_{\theta_i}(s_i,a_i) -  y_i| - \frac{1}{2}, & \textrm{otherwise}.
    \end{cases}
\end{equation*}
The algorithm backpropagtes this loss for the Q-Value map entry that represents the action $a_i$. Other entries receive a loss of zero. After processing the current experience $e_i$, the algorithm selects one experience from the replay buffer using prioritized experience replay. It uses a rank-based prioritization on the temporal difference error and a power law formula with $\alpha=2$. Recall Section \ref{sec:experience_replay} for a description of experience replay.

\citeauthor{Zeng2018} use stochastic gradient descent to train their networks. Their learning rate is set at $10^{-4}$. They use a momentum of $0.9$ and weight decay of $2^{-5}$. Their exploration strategy sets $\epsilon = 0.5$ initially and slowly lowers it to a minimum value of $0.1$ during training. They use a constant future reward discount factor of $\gamma=0.5$. They train their robot on cluttered scenes with various types of objects in both the real world and simulation in V-Rep \citep{V-Rep}. In the real world the number of objects is $n= 30$, in simulation it is $n = 10$. After the robot clears a scene, they drop a new batch of objects on the table.

The authors test their trained models in both the real-world and simulation on training scenarios and adversarial scenes, where the robot faces hand designed objects arrangements. It is important to note, that the testing procedure still performs a backpropagation step on every executed action. When an action does not lead to a change, the scene and therefore the network input do not change. Without the backpropagation step, the network would then predict the same action for the same input and get stuck in a loop. The authors use a smaller learning rate of $10^{-5}$ for this step and reset the weights of the network after every testing episode.

\citeauthor{Zeng2018} use three metrics to evaluate the performance of their model. The primary metric is the completion rate, where they define completion as successfully grasping every object in the scene. The second metric is the grasp success rate per completion, i.e., what percentage of executed grasps were successful. The last metric is action-efficiency, which the authors define as the quotient of the number of objects in the scene over the number of actions it takes to complete it.

For the V-REP simulation, the presented results show a completion rate of $100\%$ on random arrangements, with a grasp success rate of $67.78\%$ and an action efficiency of $60.9\%$. On challenging arrangements, the completion rate is $82.7\%$ with a grasp success rate of $77.2\%$ and an action efficiency of $60.1\%$.

The authors also experiment with several different configurations of their system, two of which are of interest to our work. The first configuration, "VPG-noreward", has a pushing reward function $R_p(s_t,s_{t+1}) = 0$, i.e., the robot does not receive any rewards for pushing. To evaluate it they introduce an additional metric which measures the success rate of a two step action sequence, of a push, followed by a grasp attempt. A "push-then-grasp" sequence is deemed successful, if the grasp is successful. The authors use this metric and the grasp success rate to compare the standard and the VPG-noreward configuration performance during training. The metrics are plotted over a rolling window of 200 grasp attempts. The grasp success rate of "VPG-noreward" increases slower than the standard version and the final success rate after 2500 iterations is about $10\%$ lower. The push-then-grasp success rate also increases slower for VPG-noreward and reaches a value of $\approx40\%$ after 2500 iterations compared to $\approx80\%$ for the standard configuration. The authors state that these results show that the VPG-noreward version is still able to learn effective policies without pushing rewards. While the rate of improvement is slower, the final performance only suffers slightly.

The second configuration of interest to our work is "VPG-myopic", where \citeauthor{Zeng2018} train their system with a discount factor of $\gamma = 0.2$. They state that this version improves the grasping performance quicker in the early stages of training. They compare this version to the standard configuration of $\gamma = 0.5$ on 11 challenging arrangements. The completion rate of VPG-myopic is $79.1\%$ compared to $82.7\%$ of the standard approach. The grasp success rate is also slightly lower, $77.2\%$ compared to $74.3\%$ and the action efficiency is $53.7\%$ compared to $60.1\%$. The authors argue that this indicates that the ability to plan long-term strategies, which stems from using a higher discount factor, is beneficial for the final performance.

\section{Hourglass System}\label{sec:Hourglass}

\citet{Ewerton2021} recently developed an evolution of the VPG system using an Hourglass network architecture to solve a pushing task. In their task, the robot has to push objects into a box which is located on one side of the table. They model their MDP in a similar fashion as \citet{Zeng2018}. They use $224\times224$ heightmaps as state representations, but do not capture color information and only use depth images. They also investigate augmenting the input with positional information, which we  disregard for brevity. The action space consist of pushing actions, as grasping is not considered in this task. They use the same pixel-wise definition of the action space as \citeauthor{Zeng2018} and also use $16$ action orientations. They set the length of the pushing action at $10$cm. Their reward function $R^p$ consists of two parts, a distance-based reward $R^d$ and a reward $R^{ib}$ for successfully pushing objects into the box.
\begin{equation*}
    R^p (s_t, a_t, s_{t+1}) = 
    \begin{cases}
        0, & \textrm{if an object fell off the table} \\
        R^d_t + R^{ib}_t, & \textrm{otherwise.}
    \end{cases}
\end{equation*}
The distance-based reward is 
\begin{equation*}
    R^d_t = \max(0,\, \Delta \, d_t^M), \textrm{ with}
\end{equation*}
\begin{equation*}
    \Delta \, d_t^M = d_t^M - d_{t+1}^M \quad \textrm{and } \quad d_t^M = \frac{1}{|\mathcal{O}_t|} \sum_{p \in \mathcal{O}_t} d_t(p).
\end{equation*}
$\mathcal{O}_t$ is the set of pixels of the heightmap representation that have a greater depth value than a certain threshold. The function $d_t(p)=||p-p_{\textrm{box}}||$ returns the euclidean distance of an individual pixel to the center of the target box $p_{\textrm{box}}$. Essentially, the authors give a reward if the mean distance of all object pixels to the box has decreased. They do not give negative rewards if the distance increased. The second reward function $R^{ib}_t = 10 \times N_t^{box}$, where $N_t^{box}$ is the number of objects that the robot pushed into the box.

\citeauthor{Ewerton2021} show that the VPG system performance for this task is limited by the use of image classification architectures such as DenseNet. As described in the previous section, \citeauthor{Zeng2018} use a pre-trained DenseNet architecture to process the input images and then feed the concatenated feature maps to the subsequent layer. For input images with a resolution of $640\times640$ these intermediate feature maps only have a resolution of $20\times20$. The following layers do not change the resolution of the processed images. In the end, the VPG system uses bilinear upsampling to produce the original resolution. Using this method leads to a bias of the resulting policy towards certain actions which we also illustrate in the next chapter. \citeauthor{Ewerton2021} approach this issue by changing the Q-Network architecture. They argue that the problem at hand is in fact an image-to-image translation problem for which an Hourglass architecture is more suitable. 

\cite{Newell2016} first introduced the Hourglass architecture for human pose estimation from depth images. The key feature of this architecture is its ability to process information across different scales. In human pose estimation, processing small scale information, i.e., from relatively small areas of an images, is required to identify individual body parts. Processing large scale information, i.e., across the full image, is then required to produce the final pose estimation. The capability to process both scales stems from the symmetric Hourglass module architecture, illustrated in Figure \ref{fig:Hourglass}. It consists of convolutional and max pooling layers which process the input image and produces a low resolution feature map. Before each max pooling layer, the network branches off and feeds the respective feature map to additional convolutional layers. In the subsequent upsampling process the network combines the upsampled output at every step with these branched off feature maps.

\begin{figure}[!ht]
    \centering
    \includegraphics[width=0.75\textwidth]{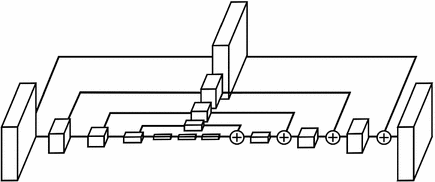}
    \caption{Hourglass module. Reprinted by permission from Springer Nature Customer Service Centre GmbH: Springer, Computer Vision – ECCV 2016, Stacked Hourglass Networks for Human Pose Estimation, \citeauthor{Newell2016} © Springer International Publishing AG 2016}
    \label{fig:Hourglass}
\end{figure}

Another problem that \citeauthor{Ewerton2021} identify is the method which the VPG system uses to extend the action space to include different action orientations. As mentioned, \citeauthor{Zeng2018} rotate the input image by 16 different rotations and feed these images into their network. The network is unaware that it is processing different representations of the same scene. This is not an issue for the VPG task, which is symmetric in the sense that it is not essential in which directions objects are pushed, as long as the objects are pushed apart. Asymmetric tasks, e.g., when it is important in which directions objects are pushed, are actually unsolvable with this method, as we further elaborate in the next chapter. \citeauthor{Ewerton2021} deal with this problem by modifying the way their Q-Network produces predictions for the different orientations. Instead of rotating the input images, the Q-Network takes just one input image with a resolution of $224\times224$ and outputs a $224\times224\times16$ feature map. Each of the $16$ channels represents a single action orientation. The Q-Network evaluates all actions for a scene in a single forward pass. As a result of this approach, the network has the ability to learn that specific orientations can lead to higher or lower rewards than others.

Additionally, \citeauthor{Ewerton2021} outline that the VPG pushing network actually has to learn two individual objectives at the same time. The first objective is predicting pushes which lead to a change in the scene. The second is determining which pushes help to solve the actual task. This duality of concerns is made clear by the VPG reward functions $R_p$ and $R_g$ which are each responsible for one of these objectives. \citeauthor{Ewerton2021} separate these concerns by introducing a second Hourglass network, called "PushMask", which they train to evaluate whether a push lead to change in the scene. They identify this problem as a binary classification problem which has to be solved for each possible action. They define the label as
\begin{equation*}
    y_i = 
    \begin{cases}
        1,& \textrm{if push lead to a detectable change} \\
        0,& \textrm{otherwise}
    \end{cases}
\end{equation*}
and train it using a binary cross entropy loss function. The PushMask outputs a binary $224x224x16$ feature map. The system combines this with the Q-Network's output using the hadamard product, multiplying the Q-Value for each action with the corresponding PushMask output value. The resulting feature map is a masked Q-Value map where the only non-zero values are those, which represent actions that the PushMask predicts to be effective. A major advantage of this approach is that a trained PushMask network can be reused for different tasks, speeding up the learning process.

\citeauthor{Ewerton2021} train their system in three stages, using a combined offline and online learning approach. The initial stage consists of training the PushMask. They first train this network using ground truth images that they generate using a heuristic. They use a Canny edge detector to find all edge pixels. They label an edge pixel $p$ as a valid push, if the depth value difference of $p$ and a pixel $p_\theta$ in the direction $\theta$ of a push is greater than a certain threshold. To improve the accuracy of their PushMask, they subsequently also use online training, executing pushes, evaluating whether they lead to a detectable change and training the PushMask on this information. After this stage, the parameters of the PushMask network are fixed and not updated any more.

The second stage consists of training the actual Q-Network. This stage starts with an offline pre-training approach. The authors generate a dataset of experiences by executing random pushes or PushMask predictions. They use each of these two alternatives with a probability of $50\%$. They then train their Q-Network with this dataset. In this stage the predictions of the Q-Network are not used to select actions for training.

In the third stage they train their combined system using a classic online reinforcement learning approach. For a given scene, they always select the action with the highest predicted Q-Value. They execute this action, observe the result and perform a backpropagation step. They do not use a target network to evaluate the expected future reward.

\citeauthor{Ewerton2021} test their PushMask individually, as well as their entire system. Here, we want to focus on their tests of the entire system. They use 100 random test scenes, each consisting of 10 objects. A test scenario ends if there are no more objects on the table or if the Q-Network predicted the same action $10$ times. In contrast to \citeauthor{Zeng2018}, they do not perform a backpropagation step during testing. They use four metrices to evaluate the performance of their models, comparing the mean values and standard deviations. The primary metric is number of objects that were successfully pushed into the box $N_{\textrm{ObjB}}$. The authors test different configurations of their system and a baseline method which uses a modified DenseNet Q-Network architecture. 
They also compare versions of their system which they trained with or without the PushMask network, as well as with or without the online reinforcement training stage. 

Their results show a clear performance improvement when using an Hourglass network architecture instead of DenseNet. Also, models trained with online reinforcement almost always outperform those that were trained without this step. The only exception are models which were given an input augmented with positional encoding. Interestingly, they achieve their two best overall results for $N_{\textrm{ObjB}}$ with a discount factor of $\gamma = 0$. Effectively, these configurations do not consider the future reward when making their decisions. The best configuration, which includes the PushMask and online training, outperforms the same configuration with $\gamma = 0.4$ by a mean value of $0.96 \, N_{\textrm{ObjB}}$. The authors state that they expected the models trained with $\gamma = 0.4$ to perform better than those trained with $\gamma = 0$. They consider the possibility that these models need more training and also acknowledge that their reward function is very informative.
\cleardoublepage


\chapter{Investigating the Long-Term Behaviour}\label{chap:investigation}
The push and grasp task investigated by \citet{Zeng2018} requires only a limited amount of foresight and a moderate discount factor. We want to investigate the behaviour of the VPG system for a task which requires long-term strategies and a large discount factor. We are interested in its capability to learn to reason about multiple actions when making its decisions. We first introduce our task setup and reward function. We describe our modifications to the original algorithm, which we made when adapting it to our problem. We then outline our training and testing procedure. Finally, we show the test results, analyze them and draw conclusions.

\section{Problem Design}
The aim when designing our task is to create a problem which requires multiple steps to solve. We choose to focus solely on pushing actions, as grasping successfully is in itself a one-step task. Our problem should therefore require a good, relatively long, sequence of pushes.

We formulate our task as a bin sorting variant. Our robot is located in front of a square workspace table. The table has an area of $0.4m \times 0.4m$ and a height of $0.2m$. We drop a predefined amount of $n=6$ objects onto the center region of the table. The center region has an area of $0.32m \times 0.15m$. The table has a border wall at each edge, preventing the robot from pushing objects off the table. We use two different object types which are distinguishable by shape and color. The first object type is a green cube with an edge length of $4cm$. The second type is a red rectangular cuboid with edge lengths of $8cm$, $4cm$ and $2cm$. We define two goal regions, one for each object type. The goal regions are the $10cm$ wide areas from the left and right edges of the table. The goal for the robot is push each of the two object types into the corresponding goal regions by pushing them. When the robot pushes one or more objects into the correct goal region we remove those objects from the scene programmatically. This simulates other robots picking up the sorted objects.

As \citet{Ewerton2021} already showed in their work, the VPG approach of rotating images to extend their action space is problematic when applied to asymmetric tasks. Asymmetric tasks like ours, require the robot to push in specific directions, dependant on the state of the environment. The problem is that two scenes, which are rotationally symmetric in at least one of the 16 rotation angles, lead to the same batch of rotated heightmaps. Figure \ref{fig:symmetry} gives an example for this situation. The network receives the same input for both scenes. Therefore, the system can not distinguish between the two scenes, which is necessary for our task. To circumvent this problem we mark the two goal regions with differently colored lines. These lines break the rotational symmetry of the scenes. These lines are also "visible" in the depth channel of the heightmap, the corresponding pixels have depth values which are slightly greater than the workspace surface. The network can now ideally learn a connection between the orientation of the marked lines and the action orientation. Figure \ref{fig:task_setup} illustrates our final task setup in the MuJoCo simulation framework.

\begin{figure}[!ht]
    \centering
    \subfigure{\includegraphics[width=0.4\textwidth]{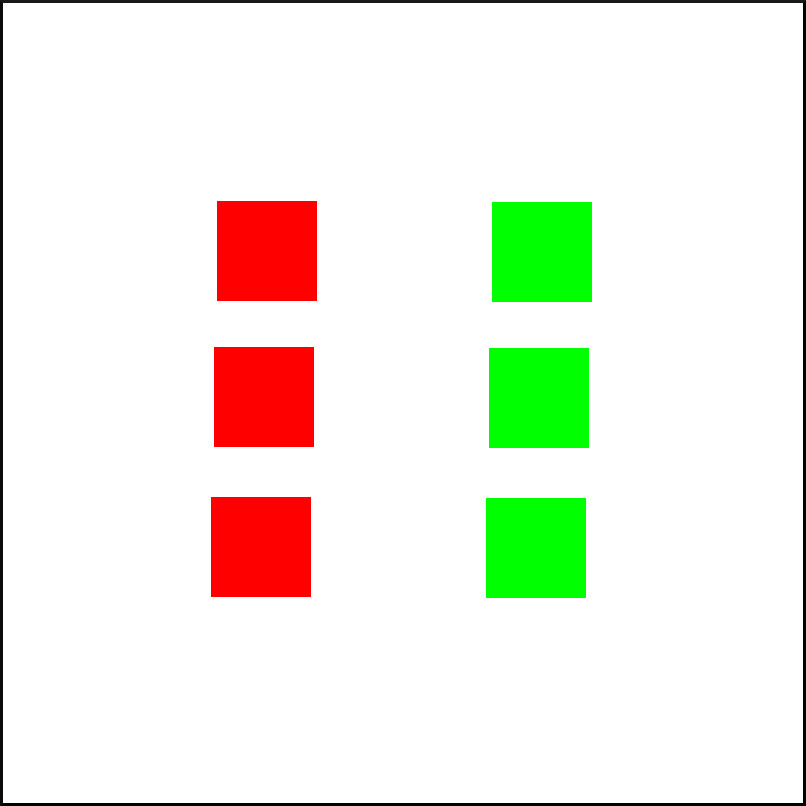}}
    \hspace{0.1\textwidth}
    \subfigure{\includegraphics[width=0.4\textwidth]{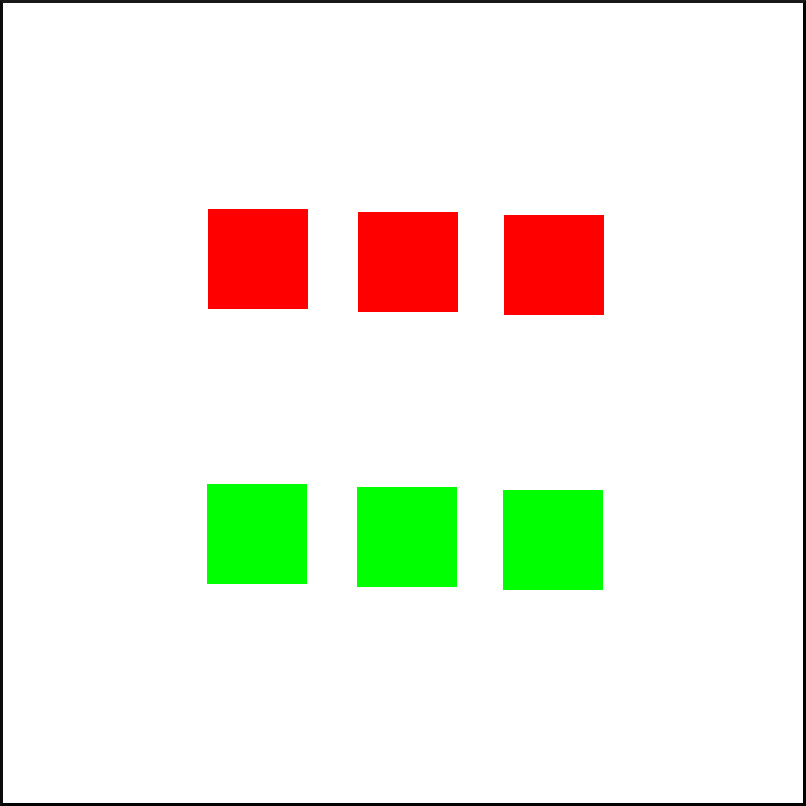}}
    \caption{Example for two symmetric scenes. These scenes lead to two equal sets of input heightmaps. The VPG system can not distinguish between the two situations.}
    \label{fig:symmetry} 
\end{figure}

We define our state and action space in a similar way as in the original VPG approach. We generate the RGB-D heightmap representation from an RGB-D point cloud capture of the Mujoco scene. Our camera is located at a fixed position above the table. The heightmaps have a resolution of $224 \times 224$. A pixel therefore represents a $3.19\textrm{mm}^2$ area of the workspace. We reduce the size of the action space, by halving the number of rotations. This decreases the amount of actions the Q-Network has to consider, while still maintaining the necessary capabilities to solve the task. We also decrease the length of the pushes from $10cm$ to $5cm$, so the robot needs more pushes to solve the task. Our terminal state space consists of the goal state $s_g$, the state in which there are no more objects left.

\begin{figure}[!ht]
    \centering
    \includegraphics[width=0.4\textwidth]{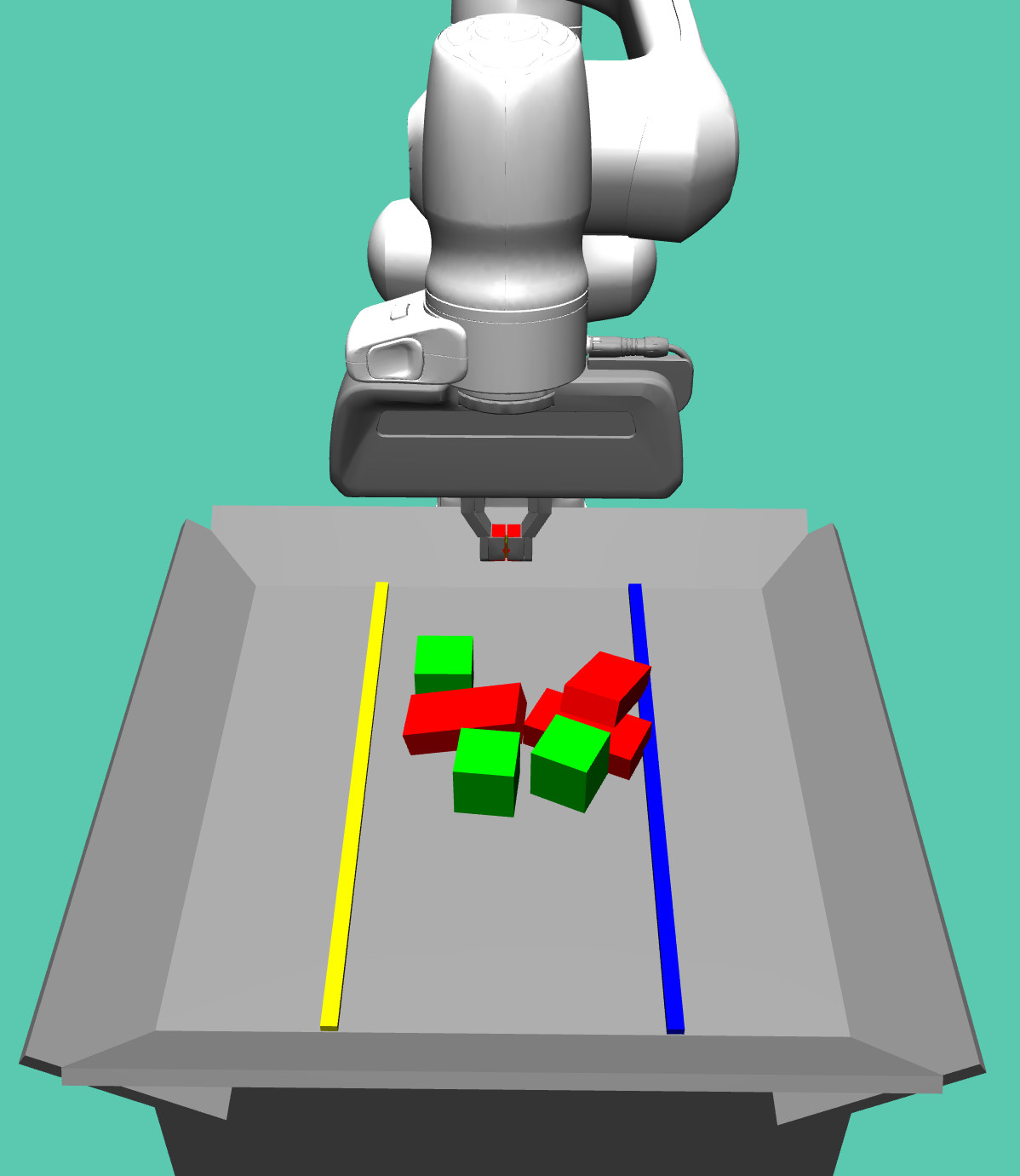}
    \caption{Task setup. The robot has to push the green cubes to the goal region marked by the yellow line and the red cuboids to the goal region marked by the blue line.}
    \label{fig:task_setup}
\end{figure}

\section{Reward Function}
For Q-Learning the reward function is the primary source of information for the system to evaluate its decisions on.
The most basic reward function rewards the system once it achieves to bring the environment to the goal state $s_g$. The first part of our reward function is therefore \begin{equation*}
    R_{\textrm{goal}}(s_t, a_t, s_{t+1}) =
    \begin{cases}
    10, & \textrm{if} \; s_{t+1} = s_g\\
    0, & \textrm{otherwise}.
    \end{cases}
\end{equation*}
This reward function alone is not sufficient to guide the robot during training. Before the robot has reached a goal state, it has no possibility to learn anything, as the reward is always zero in this case. It would have to reach a goal state solely through random exploration, which is almost impossible, especially for large amounts of objects. Therefore, we have to augment the reward function to increase the guidance for the system during training.

The reward function has to be carefully designed, so as not to misguide the robot during training. From a theoretical perspective, there are inconsistencies between the task and reward function in \citet{Zeng2018}. While the grasping reward function is suited for the goal of their task, the pushing reward function $R_p$ can cause sub-optimal behaviour: Consider a scene with a single object. The robot can choose between grasping and pushing. Compare an action sequence of a single successful grasp and a two-step action sequence of a successful push followed by a successful grasp. The first sequence has a Q-Value of $Q = R_g = 1$. The second has a Q-Value of $Q = R_p + \gamma \, R_g = 0.5 + 0.5 \times 1 = 1$. Judging from these Q-Values the two strategies are equal in terms of expected discounted reward and therefore equally optimal. Moreover, any $n$-length action sequence of $n-1$ successful pushes and one successful grasp leads to the same Q-Value, as
\begin{equation*}
    \sum_{i=0}^{n-1} \gamma^{\,i} 0.5 + \gamma^{\,n} \equiv 1 \quad \textrm{for} \; \gamma=0.5 \; \textrm{and} \; n \in \mathbb{N}^+.
\end{equation*}
Therefore, a policy which grasps the object after many successful pushes is equally optimal in terms of Q-Value as a policy which grasps the object straight away. This is counter-intuitive as a more straight-forward policy should have a higher associated Q-Value.

The intention of VPG push reward function is to guide the model towards meaningful pushes during training. Using no push rewards lead to a slower increase of the systems success rate, as their results show. Presumably, this is because the system receives no notion of a successful change without this reward. It can only learn that a push is useful when the expected future reward of the scene after the push is higher than before. This should be the case if the probability of a successful grasp increases, e.g., by separating two close objects. A push which doesn't change the scene cannot increase this probability, so the intention behind rewarding a change is correct. To retain this guidance but avoid the problem mentioned above, we decide to give the system a penalty, i.e., a negative reward, if it pushes without changing the scene. We define this part of the reward function as
\begin{equation*}
    R_{\textrm{change}}(s_t,a_t,s_{t+1}) = 
    \begin{cases}
    -0.5, & \textrm{if the push did not lead to a detectable change,}\\
    0, & \textrm{otherwise}.
    \end{cases}
\end{equation*}
With this reward function we guide the system towards effective pushes that change the scene. It avoids the aforementioned problem, because the Q-Value now reflects that a more straight-forward policy is better: Consider two policies $\pi_n$ and $\pi_m$. Let both policies consist only of effective pushes. Let them reach the goal state after $n$ and $m$ actions, respectively. Trivially, $ n < m \implies \gamma^n > \gamma^m$ for all $\gamma \in (0,1)$. Therefore, $Q_{\pi_n} = \gamma^n 10 > Q_{\pi_m} = \gamma^m 10$, i.e., the Q-Value is greater for the shorter action sequence. How much greater depends on the value for $\gamma$. For small discount factors, the difference between the Q-Values will approach zero after only a couple of steps. For large discount factors, the difference will remain noticeably greater than zero for larger number of steps.

This reward function alone is still not sufficient to guide the system towards reaching its goal. With it, the system will ideally learn that pushes that change the scene are good but receives no information of how close it is to actually completing a scene. To offer the robot some kind of notion that it is getting closer to completing a scene, we define an additional reward $R_\textrm{subgoal} = g \, \delta_\textrm{goal}$, where $\delta_{\textrm{goal}}$ is the number of objects pushed into goal position and $g > 0$ a constant factor. With this reward function, the robot receives a reward if it pushes an object into the correct goal region. The parameter $g$ determines how big the reward is. It interacts with the discount factor $\gamma$, regulating how many steps in advance the reward is noticeable in the Q-Value. For now we will set $g=2$. We can now define our complete reward function as
\begin{equation*}
    R(s_t, a_t, s_{t+1}) = R_{\textrm{goal}}(s_t, a_t, s_{t+1}) + R_\textrm{subgoal}(s_t, a_t, s_{t+1}) + R_{\textrm{change}}(s_t,a_t,s_{t+1}).
\end{equation*}

\section{Algorithm Changes}\label{sec:changes}
Building on the "Stable Baselines3" (SB3) package \citep{stable-baselines3} we adapt the VPG algorithm to the interfaces defined therein. This has the added benefit that the algorithm can be tested out on different environments, if they are defined using the OpenAi Gym interface \citep{openaigym}. The SB3 package provides a standard implementation of the DQN algorithm \citep{Mnih2015} featuring a target network, gradient clipping, experience replay buffer without prioritization, as well as configurable MLP and CNN Q-Network implementations. 

For the initial investigation of the VPG system, we extend this baseline with a prioritized experience replay buffer \citep{Schaul2016}. We also have to make some changes to the VPG system when adapting it to this package, but try to keep it as close as possible to the original system. We remove the grasping network, as our problem requires only pushes. We keep the zoom, zero-padding and rotation pre-processing step of the original approach. We normalize our RGB-D values to the range $[0;1]$. As mentioned before, we only consider eight rotations.

We replace the Batch Normalisation layers with Instance Normalisation layers in both the DenseNet towers and subsequent network layers. This approach ensures that the output for an individual input image is the same for different batch sizes. The Instance Normalisation layers have the same learnable parameters as Batch Normalization layers. These changes enable us to train the network on batches containing more than one image. We do not use pre-trained DenseNet weights.

We also change the exploration approach. The VPG exploration technique only chooses between the maximum Q-Value estimate of grasping and pushing actions. As this is not applicable here, our exploration procedure generates an action randomly out of all possible actions.

\section{Training}\label{sec:inv_train}
Our training scenes consist of three green cubes and three red cuboids. We generate a new scene for each training episode, assigning the object positions randomly in the center of the table. We also randomize the orientation of the objects. We remove any objects from the scene which fall into the correct goal region during initialization. A training episode ends if one of the following conditions is met: 1) The environment reaches the goal state. 2) The system predicts more than five consecutive actions which do not change the scene. 3) The environment does not reach the goal state in $60$ steps. We extend this final limit by $20$ steps for every object that is pushed into the correct goal region. Only the first condition corresponds to a terminal states of the MDP. The other conditions aim to limit the amount of time the system spends in an episode when it gets stuck.

We train our adaptation of the VPG system on our problem for $40.000$ steps. Each step consists of executing an action in the environment, saving the experience in the replay buffer. The replay buffer has a capacity of 2500 experiences. There is no training for the first 800 steps, during this initial stage we choose actions using exploration only. After this stage we select actions using an $\varepsilon$-greedy exploration technique. We set $\varepsilon=0.9$ at the start and lower it to a final value of $\varepsilon=0.05$ over the duration of training. Once the first 800 steps have passed, we perform a training step in each iteration of the algorithm. In the training step we use the most recent experience and sample one additional experience from the buffer. We use rank-based prioritized experience replay with a power-law distribution and $\alpha = 2$. We use a temporal difference error prioritization metric and update it for the sampled experiences after the training step. We train the network using an SGD Optimizer with a learning rate of $10^{-4}$, momentum of 0.9 and weight decay of $2 \times 10^{-5}$. We clip gradients to a maximum $2$-norm value of $10$. We train three different configurations of the network, with discount factors of $\gamma=0.5$, $\gamma=0.8$ and $\gamma=0.99$, respectively. We do not use a target network, but use the VPG approach of saving the target values when the agent makes the experience and not changing the value over the course of training.

We carry out our training on the BwUniCluster 2.0. The computation nodes we use feature Intel Xeon Gold 6230 processors and NVIDIA Tesla V100 graphics cards. Our training algorithm is sequential, i.e., requires only one processing core, and uses one graphics card. It uses about 20 GB of main memory. One training run of 40.000 iterations takes approximately 48 hours on the cluster.

\section{Testing}\label{sec:inv_test}
We test the trained model on 25 scenes that we generate in the same way as the training scenes. We also test on 10 scenes, where we randomize the distribution of object types, e.g., generating two green cubes and four red cuboids instead of the default distribution. Additionally, we test on 5 hand designed challenging arrangements, where we stack six objects closely together, or place each object in the goal region of the opposite type. See Appendix Figure \ref{app:challenging} for pictures of the challenging scenes. We follow the VPG approach of backpropagating with a smaller learning rate during testing, in order to adjust for actions which do not change the scene. The learning rate is $10^{-5}$ and we reset the weights after every testing episode.

\section{Result \& Analysis}\label{sec:inv_res}
When observing the behaviour of the models trained with $\gamma=0.5$ and $\gamma=0.8$, one notices very inconsistent performance. Sometimes, the robot pushes a couple of objects into the correct goal region, before getting stuck in scenes with two remaining objects. It then pushes these objects around aimlessly. Other times, the robot is not able to perform a single push that leads to a change in the scene, which leads to the abortion of the training episode after six timesteps. The behaviour of the model trained with $\gamma=0.99$ is even worse, as it can rarely generate a push which leads to a change in the scene.

To systematically evaluate the performance of the trained models we use three metrics. The first is the scene completion percentage, which shows how many testing scenes the model is able to solve. The second is the mean and standard deviation of the maximum amount of objects that the robot pushes into a goal region $G_{\textrm{max}}$ over all testing scenes. The third is the change percentage, i.e., the percentage of actions that lead to a change in the scene.

Table \ref{tab:inv_result} lists the test results for the trained VPG models. They show that all models are unable to solve the task except for a single test scene. The models with discount factors of $\gamma=0.5$ and $\gamma=0.8$ are only able to push two objects into the goal region on average. The model with a discount factor of $\gamma=0.99$ shows the worst performance, indicating that issues arise when using very high discount factors.

To analyze why the models are not able to learn the desired behaviour, we take a closer look at the predicted Q-Values and the type of actions the model selects in the following two sections.
\begin{table}[ht!]
\caption{Test results for the VPG models}
\vspace*{0.5cm}
\centering
\begin{tabular}{c||cccc}
\toprule
Discount factor $\gamma$ & Complete \% & $\overline{G}_\textrm{max}$ & $\sigma(G_\textrm{max})$ & Change \%\\
\midrule 
0.5 & 2.5\% & 1.9 & 1.55 & 69.1 \%\\
0.8 & 0 & 2 & 1.20 & 68.0 \% \\
0.99 & 0 & 0.075 & 0.27 & 19.0 \% \\
\bottomrule
\end{tabular}
\label{tab:inv_result}
\end{table}

\subsection{Overestimation of Q-Values}
One can compare the Q-Values predictions with the true Q-Values to evaluate how accurate they are. One calculates the true Q-Values with the reward the agent receives during testing. Figure \ref{fig:overestimation} shows the comparison of Q-Values for an example test scenarios for the model trained with $\gamma=0.5$. It shows two problems: The first problem is that the model cannot predict the reward value at timestep $t=8$ correctly, it is "surprised" by the reward it receives. In most of the other timesteps its predictions are greater than the true Q-Values. This points to the known problem of overestimating Q-Values when using a Q-Network without a target network. The behaviour of the model trained with $\gamma=0.8$ is similar. For $\gamma=0.99$ the overestimation problem actually leads to diverging Q-Values. After the first 40.000 iterations, the predicted Q-Values lie in the range $[50;55]$, which is very far from the true Q-Values.

Interestingly, one test scene does not exhibit the overestimation problem. This is the only scene that an agent, trained with $\gamma=0.5$, completed. Figure \ref{fig:underestimation} shows the Q-Values for this scenario. Here, the predictions do not overestimate the true values. They are still not ideal, as the model is not able to predict the reward it receives at timestep $t=1$. It also does not predict the reward it receives for completing the scene. We suspect the reason for this behaviour to be that the model does not see the goal state often enough during training. The training algorithm with $\gamma=0.5$ reaches the goal state in training only three times. As a result, it has not learnt that it receives a big reward for completing the scene. The question remains, why the Q-Values are more accurate in this case than in most other cases. We suspect that this behaviour is connected with another problem of the VPG approach, its bias towards certain actions.

\begin{figure}[!ht]
    \centering
    \subfigure[Overestimation]{\label{fig:overestimation}\includegraphics[width=0.9\textwidth]{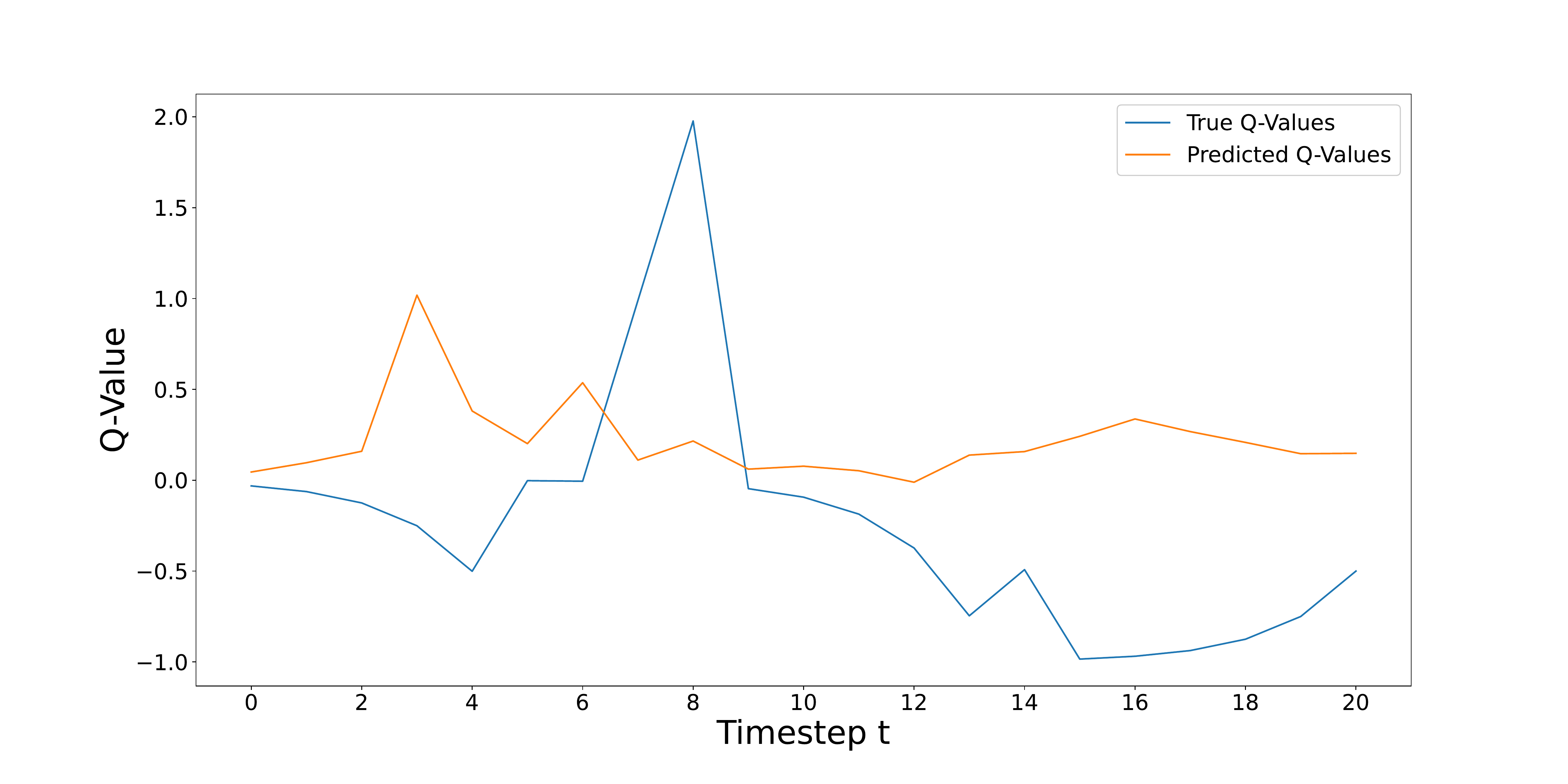}}
    \centering
    \subfigure[Underestimation]{\label{fig:underestimation}\includegraphics[width=0.9\textwidth]{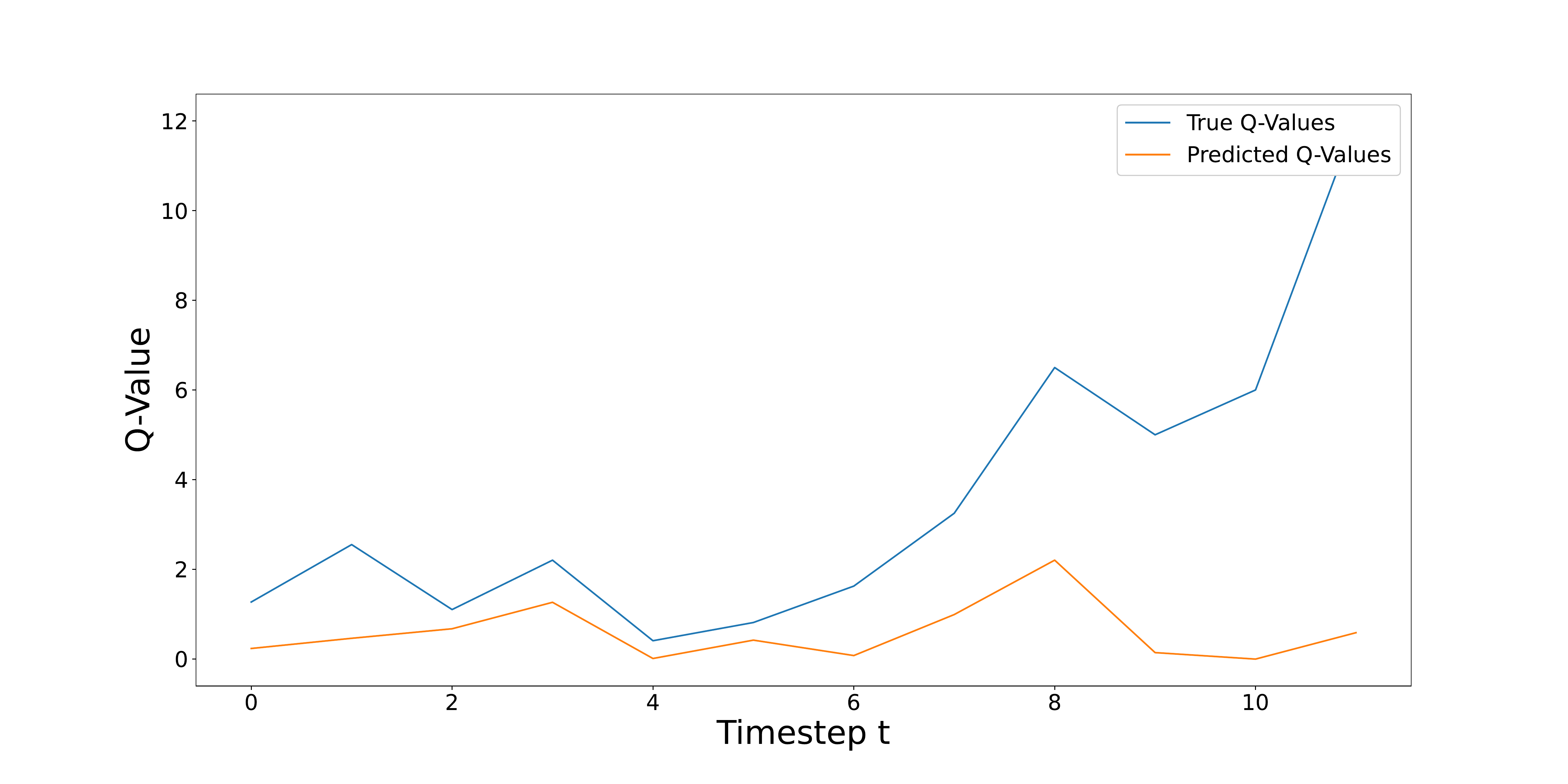}}
    \caption{Comparison of predicted and true Q-Values for $\gamma=0.5$}
    \label{fig:VPGQValues}
\end{figure}

\subsection{Interpolation Bias}\label{sec:bias}
As described in Section \ref{sec:Hourglass}, \citet{Ewerton2021} showed that the VPG system is biased towards specific actions due to its upsampling procedure. For the input image resolution of $640\times640$ the output of the Q-Network are $20\times20$ Q-Value basis points, which it then upsamples for the final prediction. \citet{Zeng2018} use bilinear interpolation for the upsampling process. This interpolation can never lead to higher Q-Values than those of the basis points. Therefore, the global maximum, i.e., the action with the highest Q-Value, will always be close to the maximum of the basis points. Figure \ref{fig:gridbias} shows which actions the agent selects during testing. The grid of basis points is clearly visible. Every executed action is represented by a pixel that is near the basis points of the interpolation. This explains the behaviour observed in the previous section. When objects lie in positions where actions represented by basis points are useful, the predictions can be accurate. But the model does not have the sufficient capacity to predict accurate Q-Values for other actions. This is a severe limitation of the action space of the system. Instead of the theoretical $224\times224$ actions per rotation, the system actually only has about $20\times20$ actions per rotation at its disposal. 

\begin{figure}[!ht]
    \centering
    \includegraphics[width=1\textwidth]{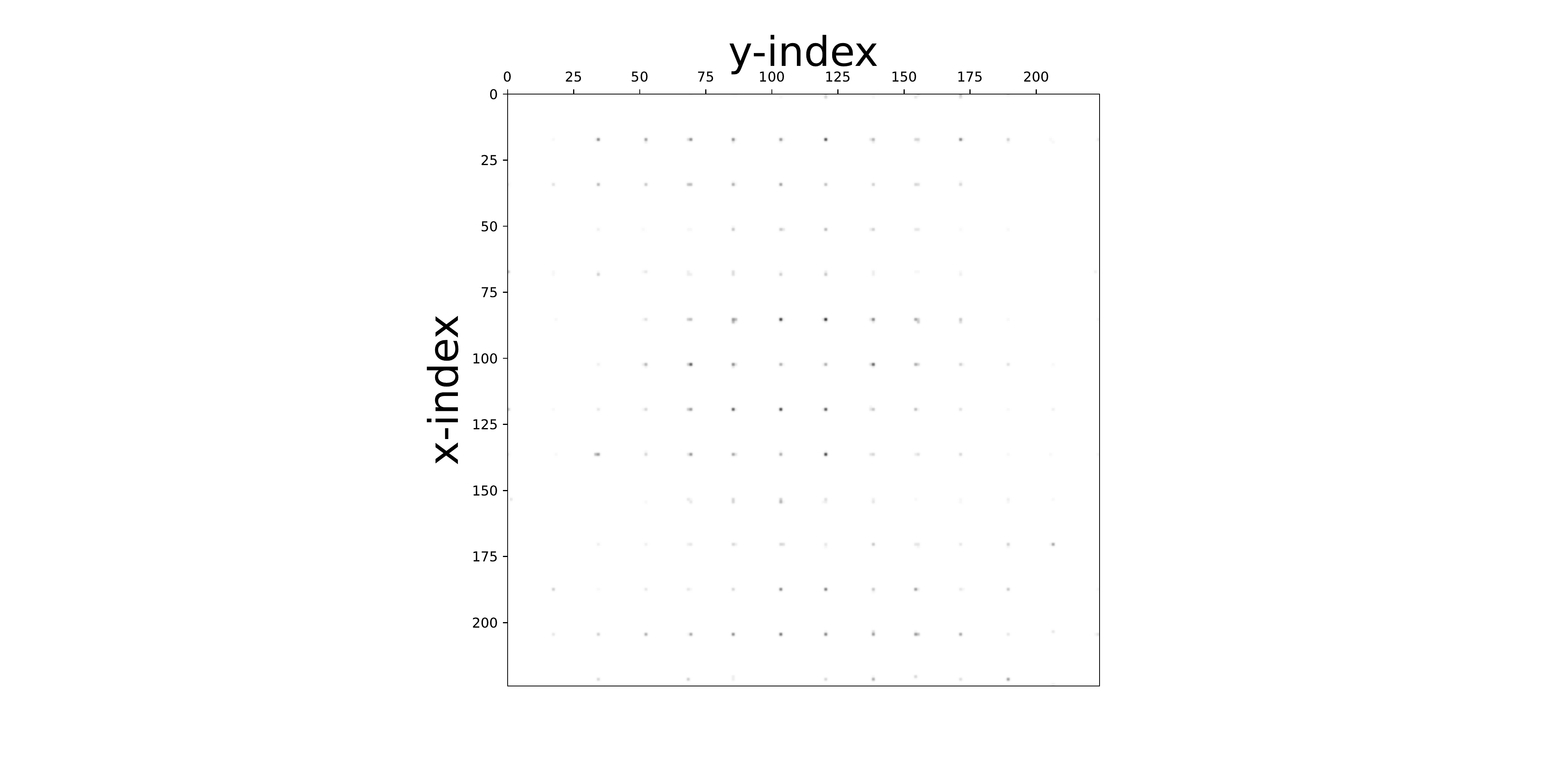}
    \caption{Heatmap showing the action occurrence by pixel index. Black points represent actions which the model selects. Basis point grid is clearly visible.}
    \label{fig:gridbias}
\end{figure}

This interpolation procedure also contributes to the overestimation of Q-Values. Consider an experience with any label $y_t > 0$ and an action $a_t$ which is not on a basis point. The prediction $Q_\theta(s_t,a_t)$ is the result of the bilinear interpolation of the basis points. Because bilinear interpolation cannot increase the global maxima of the basis points, the system can only generate the correct value for $Q_\theta(s_t,a_t)$ in two cases. Either at least one basis point must have a value $y_{b} > y_t$ or four basis points must have a value of $y_{b} = y_t$. Therefore, even if $a_t$ represents the best possible action, the predicted Q-Value of at least one basis point will be greater than or equal to the predicted value for $a_t$. Most likely, one of the basis point values will be greater than $y_t$. This behaviour leads to an overestimated Q-Value, even if $y_t$ is not itself overestimated. In later training iterations the estimated maximum future reward of this state will be greater than $y_t$, because the maximum is the overestimated value of a basis point, instead of the possibly accurate value of $a_t$. For large discount factors, e.g., $\gamma=0.99$, this almost undiscounted overestimation leads to increasing target label values, which in turn leads to increasing basis point Q-Values, resulting in an overestimation cycle. The VPG system never uses random actions in its exploration process, as they always use the maximum of either the pushing or grasping actions to select their actions. With this approach the model is trained on labels close to or on basis points only. Combined with the lower discount factor this limits the described problem in their case.

\subsection{Conclusions}

Our results show that the VPG model is not suited for our problem in its current form. The two central issues are the overestimation of Q-Values and the bias towards certain actions caused by the interpolation procedure. While the first issue could be reduced with target networks \citep{Mnih2015} or Double DQN \citep{VanHasselt2016}, the second is structural, stemming from the use of the DenseNet architecture and bilinear interpolation. Our problem requires more actions than the limited amount of basis point actions. It also requires an extended amount of foresight, represented by a large discount factor. The other problems that \citet{Ewerton2021} identified also remain. The VPG system is unable to distinguish between symmetric scenes, a problem which we tried to mitigate by marking our goal regions. In conclusion, we assert that it is not able to learn long-term policies for problems like ours. We therefore focus on the Hourglass approach of \citet{Ewerton2021} in the following work, adapt it to our problem and optimize its behaviour for our task.
\cleardoublepage


\chapter{Optimizing the Long-Term Behaviour}\label{chap:optimization}
In this chapter we detail the modifications we made to our approach in an effort to solve our task. Most importantly, the modifications include exchanging the VPG network with the Hourglass network \citep{Ewerton2021} and adapting Double Q-Learning \citep{VanHasselt2016}. We also describe discount factor scheduling \citep{Francois-Lavet2015}, whose performance impact we investigated. Additionally, we explain the upper confidence bound exploration method \citep{Auer2002} and our reasoning behind introducing it. We also touch upon our choice of loss function. We then outline our training and testing configuration and list the different hyperparameter configurations we experimented with.

\section{Hourglass Architecture}
We replace VPG Q-Network with the Hourglass network used by \citet{Ewerton2021}. Refer to Section \ref{sec:Hourglass} for a description of their system.  To be able to utilize the Hourglass network, we have to make some changes to the observation space of our environment. As the Hourglass network only uses depth images as input, we remove the color channels from the observations. The Hourglass network also removes the need of rotating the input images to generate Q-Values for different action rotations. In addition, we remove the padding and zooming pre-processing step, resulting in an input image resolution of $224\times224$. 

We also make a couple of changes to the Hourglass Q-Network. We replace the Batch Normalisation layers with Instance Normalisation for the same reasons as outlined in Section \ref{sec:changes}. We remove the last ReLu layer of the network, as this layer would constraint our Q-Values to be positive. Because we also use negative rewards, our Q-Network must have the ability to generate negative values for the corresponding actions. As we only consider eight rotations, we also decrease the number of output channels of the network to this number. 

\subsection{Mask Net}
\citet{Ewerton2021} use an additional Hourglass network in their system and train this network to predict whether an action leads to a change in the environment or not. The final layer of the mask network is a sigmoid function, producing values in $(0,1)$. They apply a threshold $\tau$ to the output, which yields a binary mask. They compute the hadamard product of the Q-Network's output and the binary mask to produce their final predictions. Therefore, the system sets the Q-Value of every predicted "non-changing" action to zero. Negative rewards and Q-values are possible in our problem, so we have to change this masking procedure. Instead of generating a binary mask, we assign the values of all non-changing actions to the largest negative float value. We set mask values for "changing" actions to zero. The modified masking procedure then sums the mask and the Q-Network output. Ideally, the Q-Values of all actions which do not produce a change are then close to the largest negative float value, while the others remain unchanged. We set the mask threshold to the same value as \citeauthor{Ewerton2021}, i.e., $\tau = 0.14$.

The user can decide whether or not to train the mask network when running our implementation. If so, we build the target labels in the same way as \citeauthor{Ewerton2021}, setting the $y_t^{\textrm{mask}}=1$ if an $a_t$ lead to a change and $y_t^{\textrm{mask}}=0$ otherwise. The user can also preload a mask network snapshot, in order to speed up training. For our experiments, we always start from scratch, without preloading and train the mask network parallel to the Q-Network. We want to evaluate the systems performance when starting from a blank slate.

\section{Double Q-Learning}
As described in Section \ref{sec:doubleQ}, Double Q-Learning is a method to reduce the overestimation of Q-Values for Q-Learning algorithms \citep{VanHasselt2016}. It decouples action selection and action evaluation when estimating the future reward. A pre-requisite of employing Double Q-Learning in DQN algorithms is the usage of a target network. When estimating the future reward to calculate the target label value in the backpropagation step, the Q-Network selects the action with the highest associated Q-Value. The target network then evaluates this action, returning its Q-Value estimate for it. This reduces the likelihood of selecting overestimated Q-Values when compared to the default approach.

As the Stable Baselines3 package already contains the methods to create and initialize a target network when creating a DQN model, we only have to modify the method which calculates the target label value. Instead of using the action with the highest associated Q-Value from the target networks output, we first perform a forward pass on the Q-Network. We then extract the index of the action with the maximum Q-Value and use it the select the corresponding value from the target network's output.

\section{Discount Factor Scheduling}
The DQN and Double DQN algorithms initialize the the target network with the same random weights as the Q-Network \citep{Mnih2015,VanHasselt2016}. Therefore, any bias that is initially present in the Q-Network will also be present in the target network. We suspect that this could slow down learning, especially if the initial random Q-Values are highly overestimated. To counteract this, we experiment with a discount factor scheduling approach \citep{Francois-Lavet2015}. The discount factor $\gamma$ determines how much of the future reward we add to the target. We set $\gamma=0$ at the start. In other words, we train the network only on the immediate rewards in the early iterations of training. With this approach, any initially overestimated Q-Values will not influence the target label in the beginning. We increase $\gamma$ iteratively until it reaches its final value after the first 10.000 iterations. We increase $\gamma$ only when we update the target network, to avoid too erratic targets. We want to investigate, whether this approach leads to an increased performance of the trained model when compared to configurations with a static $\gamma$.

\section{Exploration}
As we mentioned in Section \ref{sec:changes}, the VPG exploration procedure is not applicable in our case, since we only consider one action type. \citet{Ewerton2021} only use an exploration strategy in the first, offline, training stage. They generate experience by selecting an action randomly from all actions or from the actions predicted by the mask network, with a probability of 50\% for each possibility. In the last stage of their training, they only select the action with the highest predicted Q-Value, not using an exploration technique. We want to test the capabilities of the system using online reinforcement learning only, so we have to devise a different exploration procedure, to facilitate the exploration of the action space.

The most basic exploration strategy is using an $\varepsilon$-greedy approach, selecting a random action with a probability of $\varepsilon$. Because our action space contains a lot of actions which do not change the scene, we try to enhance the random selection of an action with a heuristic based on the depth image of a scene. We generate a binary depth image, applying a threshold of $1\textrm{cm}$ to the input depth values. We then dilute this image using max-pooling with a kernel size of $34$ and a zero-padding of $17$ pixels on each side. We trim the resulting mask to the original input resolution. The kernel size represents an area with edge lengths of $6cm$, which is a bit more than the length of our push action of $5cm$. As a result, the diluted pixels represent the area of the workspace close to or on the objects. The rest of the image represent actions which should not produce a change to the scene. Our exploration strategy selects an action from this area with a probability of $\varepsilon$.

In the early attempts of training the adapted system, we noticed a bias towards certain action orientations during the training process. To counteract this, we added an upper confidence bound (UCB) approach \citep{Auer2002} to the exploitation strategy to balance out the selection of action orientations. When applied to Q-Learning, the UCB approach selects an action with
\begin{equation*}
    a_t = \argmax_a \left( Q_\theta(s_t,a) + c \sqrt{\frac{\textrm{log}\, t}{N_t(a)}} \right),
\end{equation*}
where $c$ is a confidence value and $N_t(a)$ is the number of times the algorithm selected the action $a$. Essentially, the UCB approach explores the action space by adding an "uncertainty" value to actions which were rarely selected in the training process. We apply this approach to our problem by defining the action subspaces $A_{1-8}$ which each represent an individual action orientation. We then select actions by first extracting the action with the highest Q-Value from each subspace, i.e.,
\begin{equation*}
    h_i = \argmax_{a \in A_i} Q_\theta(s_t,a).
\end{equation*}
Then we add the uncertainty term to the Q-Value of each $h_i$ and select the maximum value from this sum with
\begin{equation*}
    a_t = \argmax_{h_i} \left(Q_\theta(s_t,h_i) + c \sqrt{\frac{\textrm{log}\, t}{N_t(A_i)}}  \right).
\end{equation*}
$N_t(A_i)$ represents the amount of times the algorithm selected an action from the respective subspace. If one or more action orientations are rarely chosen, their uncertainty value will increase, eventually leading to their selection.

In summary, our exploration strategy is a combination of an $\varepsilon$-greedy and UCB approach. With a probability of $\varepsilon$ we select a random actions from the area around or on the objects. In other cases, we select an action using the adapted UCB algorithm. It is worth to note, that we increase the counter $N_t(A_i)$ only in the latter case. The confidence parameter $c$ defines how much the uncertainty term influences the selection. Setting $c=0$ is equivalent selecting an action without adding an uncertainty term, i.e., not using UCB. We want to investigate if the employment of UCB has a positive effect on the final performance of the models.

\section{Loss Function}\label{sec:opt_loss}
The VPG and Hourglass systems use different loss functions in their training process. \citet{Zeng2018} rely on the Huber loss function while \citet{Ewerton2021} use a mean squared error (MSE) loss. Both loss functions produce the same output, if the difference between the prediction and the target is small. If the inputs diverge, the Huber loss function produces a linear loss value, while MSE loss remains quadratic in the input difference. Therefore, the Huber loss function is less sensitive to outliers \citep{Huber1964}. We want to investigate how much the choice of the loss function influences the performance of the system. The reward for completing the scene is large compared to the rewards during the steps leading up to that point. We suspect that MSE works better for this problem, as the training process should be sensitive to these outliers.

\section{Training \& Testing}
Our training scenes are the same as described in Section \ref{sec:inv_train}. The configuration of the training parameters also largely remains the same. The changed parameters include the batch size, the optimizer learning rate and the $\epsilon$-schedule for our exploration procedure. We now use a batch size of 15, which is possible because we reduced the amount and the size of the input images by removing the zooming and rotating pre-processing steps. The networks input of a single experience now only consists of a single-channel $224x224$ image and not eight 4-channel $640x640$ images. We increase the learning rate of the SGD optimizer to $10^{-3}$. We keep the momentum at 0.9 at the weight decay at $2 \times 10^{-5}$. For the mask network we use an Adam optimizer, following \citeauthor{Ewerton2021}'s approach. We also increase its learning rate to $10^{-3}$. Our exploration schedule now decreases $\epsilon$ from $\epsilon = 0.9$ to $\epsilon=0.05$ over the first 20.000 iterations. During the last 20.000 steps we keep $\epsilon$ constant at this value. 

In the training method, we backpropagate on the Q-Network first, and afterwards on the mask network. We use same experience samples for both networks. For Double Q-Learning it is worth to mention, that we only use a masked Q-Network output to select the estimated best action for the subsequent state of an experience. We do not use the mask network for the evaluation of that action with the target network, or the computation of the current Q-Value prediction for the action of the experience. With this approach, we prevent backpropagating on masked Q-Values.

We train our models on the same BwUniCluster machines as before. Refer to Section \ref{sec:inv_train} for their description. Because 40.000 training iterations now take a couple of hours longer than 48 hours, the job time limit of the cluster, we split the training process into two parts. After 20.000 iterations, we save the model and replay buffer and continue training in a new job.

We test the models on the same test set we used for the VPG models, as described in Section \ref{sec:inv_test}. During testing we follow the approach in \citet{Zeng2018} and perform a backpropagation step after every executed action. We increase the learning rate for this step by the same factor as the training learning rate, resulting in a value of $10^{-4}$. We reset the weights of the model after every testing episode.

\subsection{Parameters}
To find the best configuration, we train 20 models with different values for the hyperparameters of interest to our work. The configurations differ in four parameters: 1) The (final) discount factor $\gamma$ 2) The discount factor scheduling approach 3) the loss function $\mathcal{L}$ 4) the UCB confidence factor $c$. Table \ref{tab:params} shows the different possibilities for these parameters. We keep the other parameters equal for all models. We describe and analyze the behaviour of the trained models in the next chapter.
\begin{table}[ht!]
\caption{Parameter values}
\vspace*{0.5cm}
\centering
\begin{tabular}{ll}
\toprule
Parameter & Values \\
\midrule 
(Final) discount factor $\gamma$ & 0; 0.8; 0.99\\
$\gamma$-schedule & Static $\gamma$ ; $0\rightarrow\gamma$ over the first 10.000 iterations \\
Loss function $\mathcal{L}$ & MSE; Huber \\
UCB confidence $c$ & 0; 1 \\
\bottomrule
\end{tabular}
\label{tab:params}
\end{table}

\cleardoublepage



\chapter{Evaluation}

In this chapter we evaluate the behaviour and performance of the trained models. First, we examine whether they can solve our task reliably. Additionally, we investigate the accuracy of the Q-Value predictions and if they show any similar limitations like the interpolation bias of the VPG models. We also show which parameter configuration works best. The most important choice of parameter is the discount factor. If the models are able to learn with a very high discount factor, it would indicate that they posses the necessary capabilities to learn long-term strategies for other tasks as well.

\section{Results}
When looking at the behaviour of the optimized models, one can clearly see a large improvement from the original VPG approach. Almost every model trained with $\gamma > 0$ is able to solve the task consistently. We introduce another metric: the mean and standard deviation of the amount of actions $N_A$ that the model needs to complete a scene over all completed scenes. A lower score in this metric represents that the model can solve the task quicker.

Table \ref{tab:opt_results} shows the results for selected models over all testing scenarios. See Appendix Table \ref{app:results} for our full results. The best performing model $\textrm{M}_\textrm{best}$  uses $\gamma=0.99$, no discount factor scheduling, an UCB confidence value of $c=1$ and the Huber loss function. The second best model $\textrm{M}_{2^\textrm{nd}}$, with comparable test scores, uses $\gamma=0.99$, a $\gamma$-schedule, no UCB and the Huber loss function. We also show the best performing model out of those trained with $\gamma=0$, $\textrm{M}_{\gamma=0}$. This model uses the Huber loss function and no UCB exploration. These results show that a high discount factor is essential for solving the problem. A policy which only considers the most immediate action cannot reliably reach the goal state in our task.

\begin{table}[ht!]
\caption{Test results for the models}
\vspace*{0.5cm}
\centering
\begin{tabular}{c||cccccc}
\toprule
Model & Complete \% & $\overline{G}_\textrm{max}$ & $\sigma(G_\textrm{max})$ & Change \% & $\overline{N}_A$ & $\sigma(N_A)$\\
\midrule 
$\textrm{M}_\textrm{best}$ & 95 \% & 5.975 & 0.16 & 95.5 \% & 22.9 & 9.9 \\
$\textrm{M}_{2^\textrm{nd}}$ & 95 \% & 5.95 & 0.22 & 92.6 \% & 25.2 & 12.3 \\
$\textrm{M}_{\gamma=0}$ & 32.5\% & 4.6 & 1.52 & 81.8 \% & 78.2 & 34.5 \\
\bottomrule
\end{tabular}
\label{tab:opt_results}
\end{table}

\subsection{Q-Value Predictions}
In Figure \ref{fig:MbestQValues} we show the comparison between the predicted and true Q-Values for two example test scenarios for the $\textrm{M}_\textrm{best}$ model. For the first example, the predictions are mostly accurate. The model is able to predict the final goal reward. In earlier timesteps and especially at the beginning of the episode the model is not able to predict the actual Q-Value, but underestimates it. This problem is even more noticeable in the second example, where the model consistently underestimates the true Q-Value. While the predictions are apparently good enough to solve the problem, the comparison shows that there is still some room for improvement for the accuracy of the Q-Values. For $\gamma=0.99$ the Q-Value is very high in the beginning of an episode as it reflects the accumulated reward over all steps in an episode. We suspect that the underestimated Q-Values are most likely the result of premature training termination. Double Q-Learning slows down the increase of future reward estimations in an effort to reduce overestimation. With more training iterations, the future reward estimations should increase further and the model could learn to predict the initial Q-Values more accurately. Another reason for the underestimated Q-Values could be an uncertainty of the model about whether it can actually complete the scene. The Q-Values are only high for episodes where the robot reaches the goal state. The lower initial Q-Values could therefore indicate that the confidence in success is not as high initially, but grows once the robot pushes a couple of objects into the goal regions.

\begin{figure}[!ht]
    \centering
    \subfigure[Example 1]{\label{fig:MbestQmin}\includegraphics[width=0.9\textwidth]{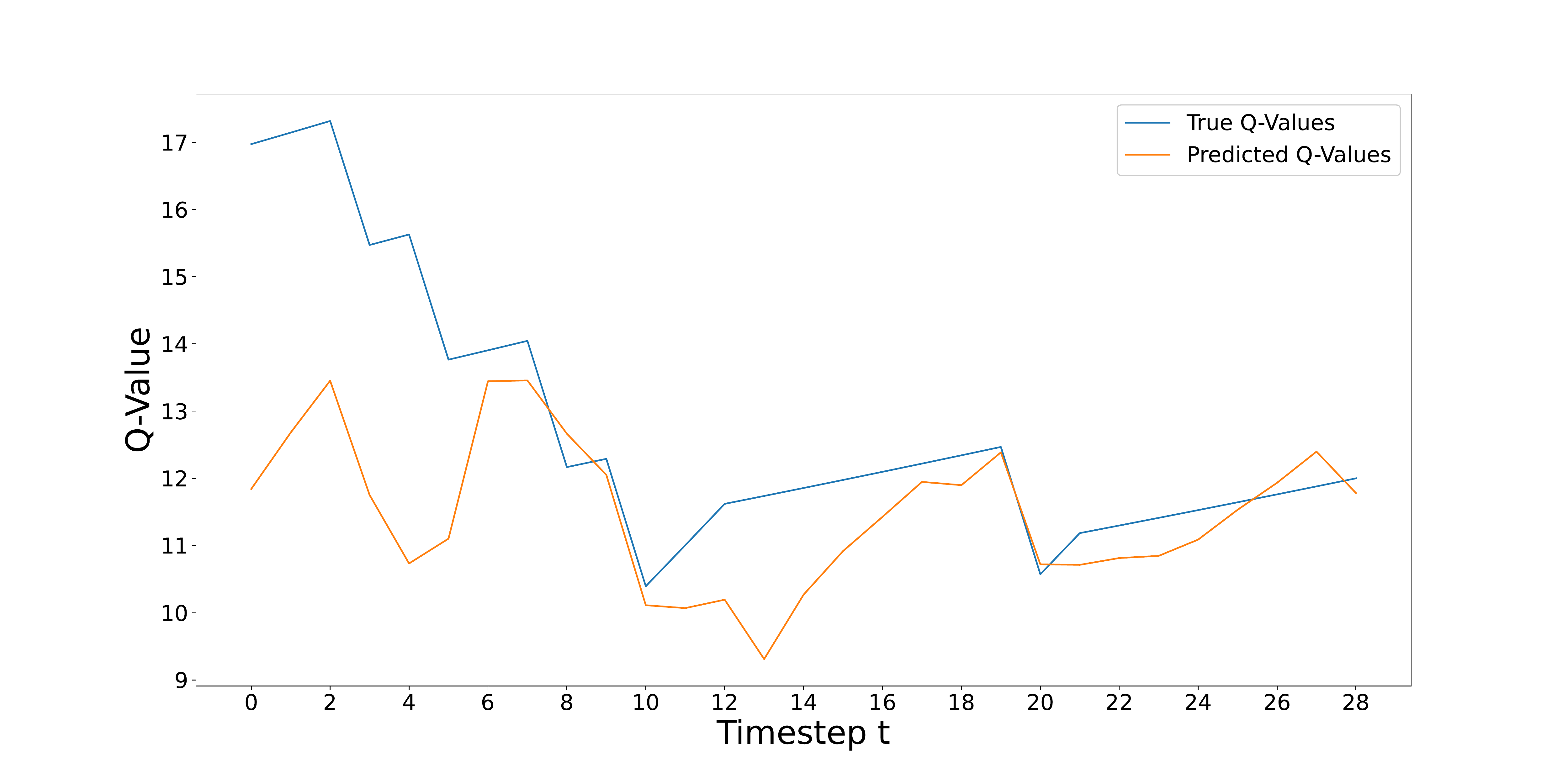}}
    \centering
    \subfigure[Example 2]{\label{fig:MbestQmax}\includegraphics[width=0.9\textwidth]{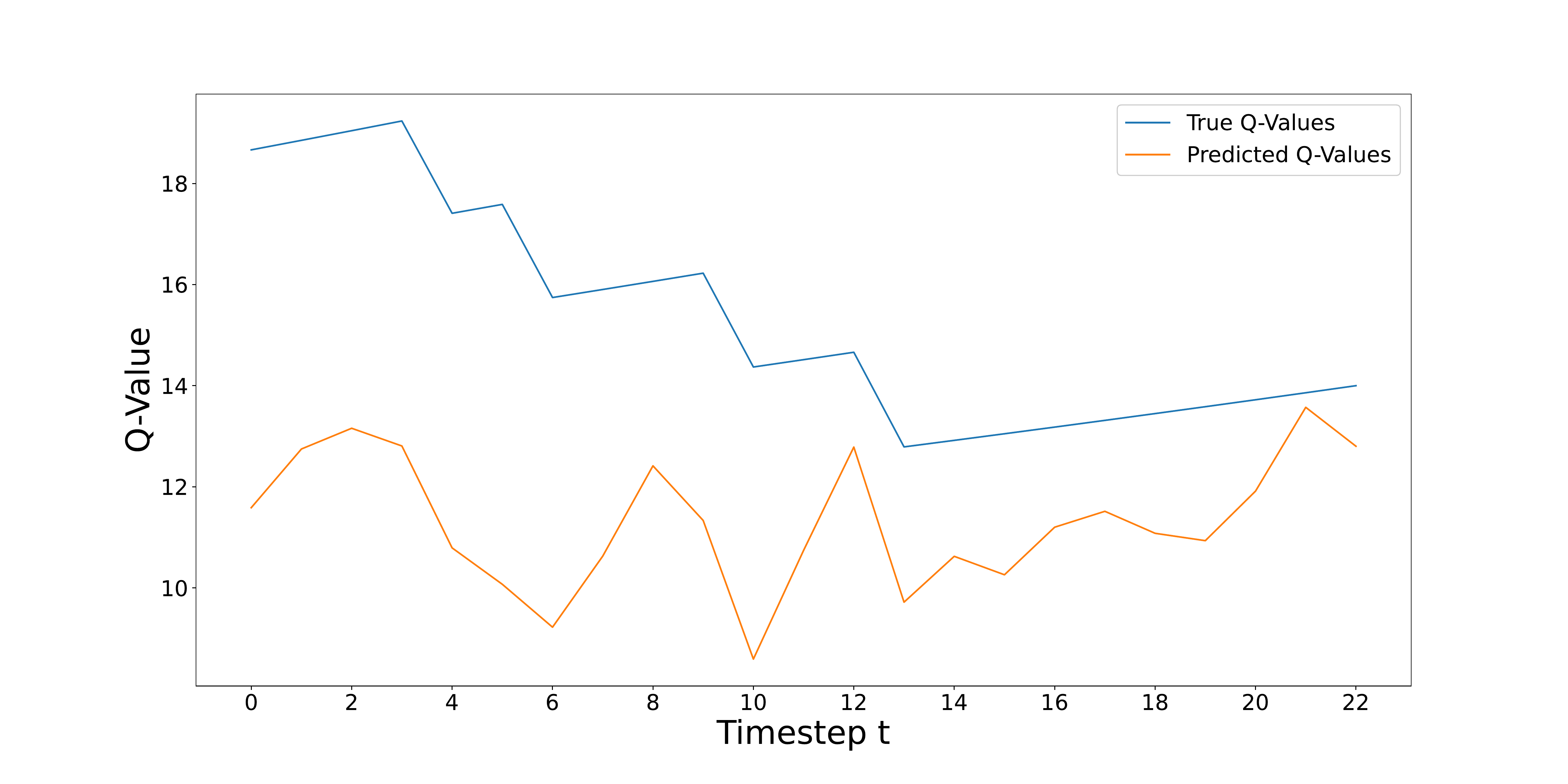}}   
    \caption{Comparison of predicted and true Q-Values for $\textrm{M}_\textrm{best}$}
    \label{fig:MbestQValues}
\end{figure}

\subsection{Action Selection}
As we described in Section \ref{sec:bias}, the upsampling approach severely limits the type of actions that VPG system has at its disposal. We examine whether the Hourglass models exhibit any similar limitations. We show the distribution of actions for $\textrm{M}_\textrm{best}$ in Figure \ref{fig:MBestHeat}. It shows that the Hourglass model can predict a larger variety of actions. There seem to be no underlying structural limitation in the capacity of the Hourglass models to predict different actions. 

\begin{figure}[!ht]
    \centering
    \includegraphics[width=1\textwidth]{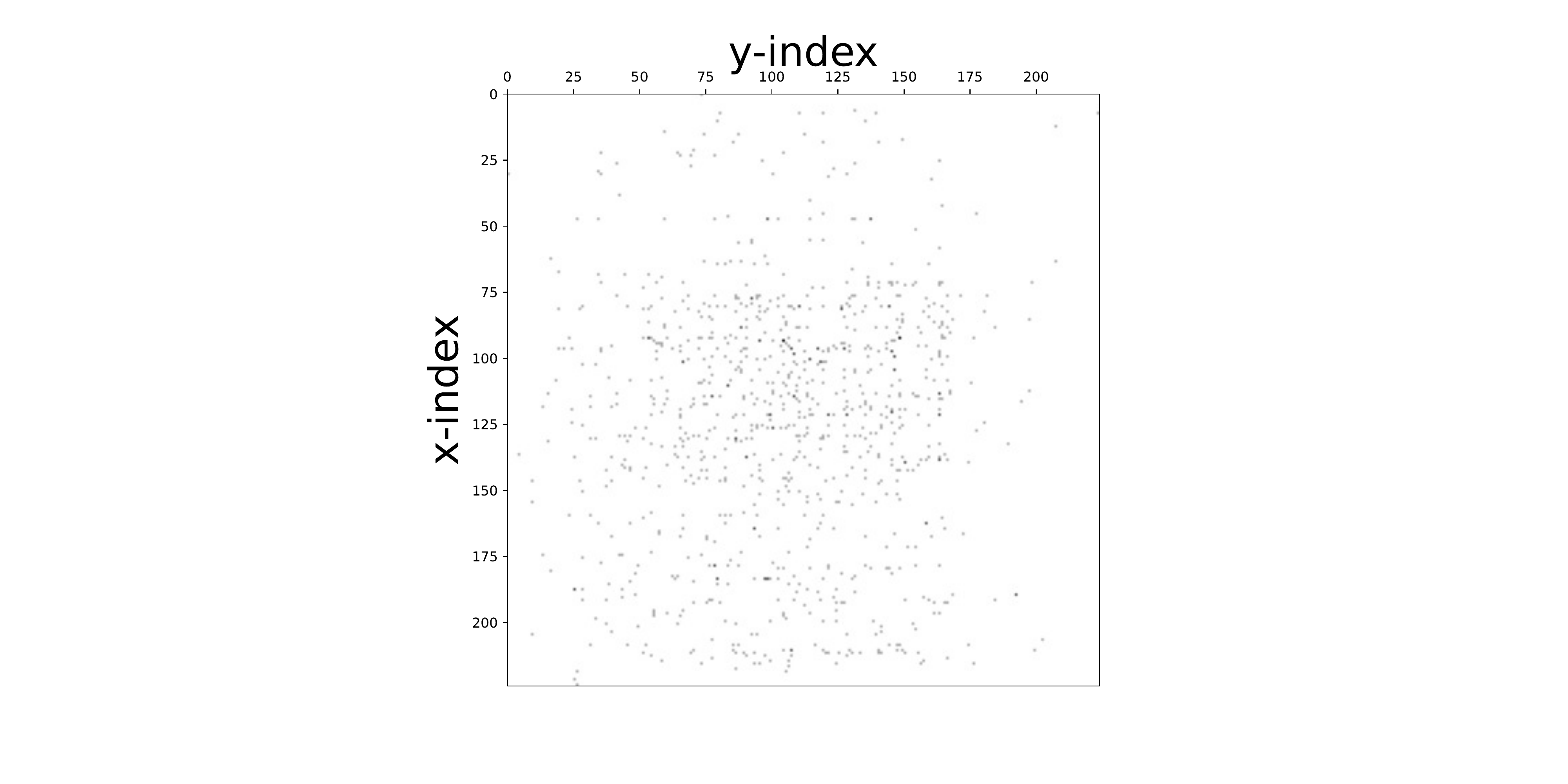}
    \caption{Heatmap showing the action occurrence by pixel index. Black points represent actions which $\textrm{M}_\textrm{best}$ selected during testing.}
    \label{fig:MBestHeat}
\end{figure}

\section{Different Parameters}
To evaluate which hyperparameters are influential for the final performance of the model, we compare the $\textrm{M}_\textrm{best}$ configuration with each configuration which is equal in all but in one hyperparameter. We also compare the test scores of the other configuration pairs which differ in exactly one parameter. Because of time constraints we were not able to perform multiple training runs for every configuration which would allow for more confidence in the following results. Confirming the observed tendencies remains a topic for future work.

\subsection{Discount Factor}
The results shown in Table \ref{tab:opt_results} already demonstrate that we need to train the model with a discount factor $\gamma > 0$ to reliably solve the task after the allocated training time. Models which use a discount factor of $\gamma=0$ only take the most immediate action into account. They do not have the capacity to plan long-term strategies as there is no information about long-term reward given to the model. For \citeauthor{Ewerton2021}'s task one-step strategies suffice because the reward function is informative enough. If the reward is sparse, like in our case, the model needs a higher discount factor.

We want to evaluate what impact the choice between $\gamma = 0.99$ and $\gamma = 0.8$ has on the performance of the model. For a theoretical comparison consider the inequality $\gamma^{\,n} \, 10 < 0.1$ which exemplifies the amount of steps $n$ after which the goal reward will be negligible because of discounting. For $\gamma = 0.8 \Rightarrow n > 20$, while for $\gamma = 0.99 \Rightarrow n > 458$. In other words, for $\gamma=0.99$ the future reward is discounted much slower than for $\gamma=0.8$. Depending on the task and reward function, a model trained with the higher discount factor can therefore take longer action sequences into consideration.

In Table \ref{tab:opt_gamma} we compare the $\textrm{M}_\textrm{best}$ model with the model $\textrm{M}_{\gamma=0.8}$ which we trained with the same parameters as the former, with the exception of setting $\gamma=0.8$. The $\textrm{M}_\textrm{best}$ model outperforms $\textrm{M}_{\gamma=0.8}$ by a small completion rate margin of $5\%$. The change percentage indicates that it has learnt to predict changing actions more accurately than $\textrm{M}_{\gamma=0.8}$. It is also able to complete scenes in a smaller amount of actions on average. When comparing the other configuration pairs which differ in the discount factor those with $\gamma=0.99$ perform equally or better than their respective counterparts in all but one case. With the limitation of one training run for each configuration, these results seem to indicate that using a higher discount factor of $\gamma = 0.99$ is beneficial for the final performance on our task. The performance difference isn't large, because the best models can solve our scenes in about 20-30 steps. We suspect that the difference would be greater for more complex problems, for which the agent needs to evaluate even longer strategies. With longer paths to goals and rewards the agent would have to be more reliant on making its decision based on the future reward estimations.

\begin{table}[ht!]
\caption{Test results comparison for different $\gamma$}
\vspace*{0.5cm}
\centering
\begin{tabular}{c||cccccc}
\toprule
Model & Complete \% & $\overline{G}_\textrm{max}$ & $\sigma(G_\textrm{max})$ & Change \% & $\overline{N}_A$ & $\sigma(N_A)$\\
\midrule 
$\textrm{M}_\textrm{best}$ & 95 \% & 5.975 & 0.16 & 95.5 \% & 22.9 & 9.9 \\
$\textrm{M}_{\gamma=0.8}$ & 90 \% & 5.65 & 1.25 & 87.4 \% & 29 & 16.1 \\
\bottomrule
\end{tabular}
\label{tab:opt_gamma}
\end{table}

\subsection{Discount Factor Schedule}

We experimented with a discount factor schedule to see whether it helps to stabilize training in the early stages. From our results we can see no indication that this scheduling approach is necessary. Comparing the configuration pairs which differ in the scheduling approach shows no clear advantage for either configuration. The learning curves also don't seem to be heavily influenced by scheduling, as illustrated in Figure \ref{fig:comp_gamma_frac}. While scheduling doesn't appear to hinder learning, we argue based on our results that discount factor scheduling is not required to learn long-term strategies from scratch.

\begin{figure}[!ht]
    \centering
    \includegraphics[width=1\textwidth]{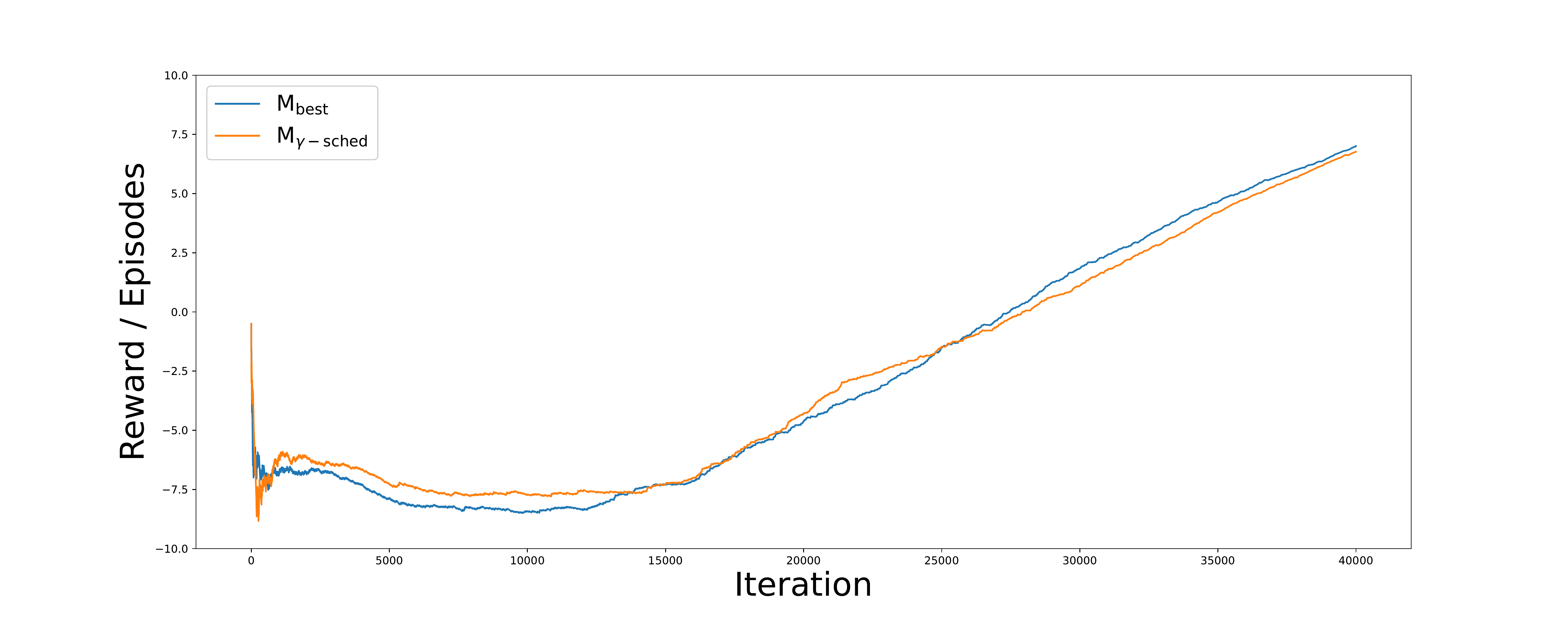}
    \caption{Comparison of learning curves for $\textrm{M}_\textrm{best}$ and the corresponding configuration with discount factor scheduling, $\textrm{M}_{\gamma-\textrm{sched}}$.}
    \label{fig:comp_gamma_frac}
\end{figure}

\subsection{Loss Function}

The choice of loss function also seems to have no impact on the final performance of the trained models. The $\textrm{M}_\textrm{best}$ model uses the Huber loss function but models trained with an MSE loss function can also achieve comparable scores. This disproves our suspicion formulated in Section \ref{sec:opt_loss}, that the model trains better if the loss function is more sensitive to outliers.

\subsection{Upper Confidence Bound Exploration}

The $\textrm{M}_\textrm{best}$ model uses the UCB exploration technique, while the model $\textrm{M}_{2^\textrm{nd}}$ does not. When comparing the individual configurations pairs which only vary in the usage of UCB, there seems to be no clear advantage in performance for either. The results seem to indicate that our adaptation of UCB has no impact on the final performance of the model. We introduced UCB because we wanted to balance the selection of different action orientations during training. But, our problem mainly needs only two push directions, towards the goal regions. In testing $\textrm{M}_\textrm{best}$ chooses these directions $88.8 \%$ of the time and $\textrm{M}_{2^\textrm{nd}}$ selects them $89.2\%$ of the time. This behaviour is understandable, as the shortest path to goals is pushing the objects straight towards the correct goal regions. The other actions are only auxiliary actions which the agent rarely chooses. We suspect that problems which require a larger variety of pushing directions and other types actions would warrant further investigation of the UCB exploration technique.

\section{Training without Target Network}

To evaluate the necessity of Double Q-Learning when using high discount factors, we trained two additional configurations using a discount factor of $\gamma = 0.99$. These models use no discount factor scheduling, no UCB exploration and the Huber loss function. The first configuration uses no target network, but follows \citeauthor{Zeng2018}'s original approach of saving future reward estimations when the agent makes an experience. The second configuration uses a target network, but no Double Q-Learning. It selects the future reward estimate by extracting the maximum Q-Value from the Q-Networks output for the subsequent state of an experience.

Both of these configurations lead to diverging Q-Values and are therefore unusable. This demonstrates that dealing with the overestimation problem is a pre-requisite for using high discount factors. It underlines the need to employ a target network and Double Q-Learning in Dep algorithms. 
\cleardoublepage


\chapter{Conclusion and Future Work}

In this chapter we summarize our findings and formulate our conclusions. We then outline possible topics for future work.

\section{Conclusion}
Creating artificial intelligence is an intriguing challenge of today's time. Teaching machines to think and reason would lay the foundation for advances in almost every scientific field imaginable. In our research, we work on Robot Reinforcement Learning, where we aim to train a robot to execute a sequence of good actions in its environment. The robot learns about good and bad actions through experience, i.e., executing actions and receiving rewards or penalties. To be able to refer to the robots behaviour as "intelligent" it needs to be able to balance short-term and long-term reasoning in an effort to collect the maximum reward possible.

Recently, researchers approached problems in the field of Robot Reinforcement Learning by using neural networks, which were previously successful in Computer Vision tasks, to process the observations of the environment. The VPG system developed by \citet{Zeng2018} is an example for such a system. It builds on a DenseNet architecture to process RGB-D images of the workspace. As a Deep Q-Learning variant the system learns to predict the action Q-Values for a given state. Deep Q-Learning is usually challenging for robotic problems because of the high-dimensional continuous joint angle action space. Instead of trying to learn to predict actions in this continuous action space, the authors discretize it dynamically. They relate action locations to pixel positions and depth information of the environment observations. Dependent on the current state of the environment, the agent has a different set of actions at its disposal. To extend their action space for different orientations, they rotate the input images in 16 rotation angles and feed them to the network individually. Each individual image represents on action orientation. In their work, \citet{Zeng2018} focus on demonstrating that two Q-Networks for two action types can be trained together. Their task only requires a limited amount of planning, which is made apparent by the comparatively small discount factor they use.

\citet{Ewerton2021} identified a couple of structural limitations of the VPG system and published their own Hourglass system in an effort to remove them. These limitation stem from the DenseNet architecture which produces low resolution intermediate feature maps. \citet{Zeng2018} use bilinear upsampling to recreate the original resolution which severely limits the amount of actions the system can actually use. Additionally, the method which the VPG system uses to extend its action space for different orientations is problematic when applied to asymmetric problems. The VPG system is not aware that it is judging different action orientations for the same scene. \citet{Ewerton2021} exchanged the DenseNet Q-Network architecture with an Hourglass architecture, which \citet{Newell2016} originally introduced for human pose estimation. They also removed the need for rotated input images as their system evaluates all rotations at once. Furthermore, they separated the concerns of identifying which actions lead to a change in the scene and predicting Q-Values by introducing a mask network. In their work, they show that their system is able to solve a pushing task far better than the VPG system. Their problem of pushing objects from a table into a box is a task which requires some amount of planning and foresight of the agent. Surprisingly, their best performing configuration uses a discount factor of $\gamma=0$, which means that the agent only considers the most immediate action when making its decisions. The authors suggest the informative reward function, which gives a distance-based reward for every action, as a possible reason for this behaviour.

In our work we investigated the suitability of the VPG system for another pushing task which requires long-term planning. We designed our problem and reward function with the requirement that the agent has to learn to accurately estimate long-term future rewards in order to solve it. We chose a bin sorting task with two object types which have to be separated into two goal regions. We only give rewards if the robot reaches a subgoal, i.e., pushes an object into the correct region, or if it reaches the final goal state. We implemented the task in a MuJuCo \citep{mujoco} simulation environment following the OpenAi gym interface \citep{openaigym} and adapted the VPG system using the Stable-Baselines3 package \citep{stable-baselines3}. We tried to keep our adaptation of the VPG system as close as possible to the original implementation. We trained the adaptation on our task for 40.000 iterations. We tested the trained models on a test set, composed of training scenes, scenes with random object type distribution, and hand-designed challenging arrangements. Our results show that the VPG system is not suited to solve tasks which require long-term action planning. In addition to the limitation caused by interpolated Q-Values, it suffers from the known DQN problem of overestimated Q-Values. For a discount factor of $\gamma=0.99$ this overestimation problem leads to diverging Q-Values and unusable model predictions. But, even for lower discount factors, the VPG models are not able to solve our task. 

We then made several changes to our approach in an effort to solve our task. These changes included replacing the VPG Q-Network with \citeauthor{Ewerton2021}'s Hourglass Q-Network and also adding their mask network. We extended their system with a target network and the Double Q-Learning technique \citep{VanHasselt2016}, again building on the Stable-Baselines3 package. In contrary to \citet{Ewerton2021}, we trained the models using online reinforcement learning only. Our main focus was training with different discount factors, to see what influence this parameter has on the performance of the models. We also investigated extending an $\varepsilon$-greedy exploration technique with a UCB exploration approach \citep{Auer2002}. Furthermore, we trained models using discount factor scheduling \citep{Francois-Lavet2015} and different loss functions, to investigate what impact they have on the models performance.

Our results show that our adaptation of the Hourglass system is able to solve the task reliably when trained with high discount factors. The models trained with $\gamma=0.99$ show the best performance, indicating that this choice of discount factor is beneficial for our problem. Our best performing model was able to complete test scenarios 95\% of the time. The best completion score of all models trained with $\gamma=0$ is only 32.5\%, showcasing the impact of the agent's ability to apply long-term reasoning on its performance. The Q-Value predictions of the best models are accurate enough to solve the task, but it is interesting to note that they tend to underestimate the true Q-Values, especially early in an episode. More training might lead to an improvement in the accuracy of these values. We could see no structural limitation of actions in the Hourglass models, like we saw in the VPG models. Our results also show that using discount factor scheduling and UCB exploration provided no benefit for training the models for our problem. The choice between a Huber and MSE loss function also did not impact the performance of the models.

Neither \citet{Zeng2018} nor \citet{Ewerton2021} use a target network in their training algorithms. \citeauthor{Zeng2018} use a moderate discount factor of $\gamma=0.5$ for a task which does not require long-term planning. \citeauthor{Ewerton2021} attain the best performance using $\gamma=0$, which means that the model disregards the future reward and only takes the most immediate action into consideration when it makes its decisions. With our task we demonstrate that one needs high discount factors if the information of the reward function is limited. Furthermore, we show that Double Q-Learning is a requirement for using these high discount factors. Without this method, training with $\gamma=0.99$ leads to diverging Q-Values for both networks. 

In conclusion, our work shows that the Hourglass model is able to plan long-term strategies when combined with the Double Q-Learning technique. Even with a very high discount factor of $\gamma=0.99$, it can predict a sequence of good actions to a satisfying degree.

\section{Future Work}

In addition to repeating our experiments to gain more confidence in our results, we also plan to increase the difficulty of the task. Our best performing models can solve our task in 20 to 30 steps. We want to investigate the behaviour for tasks which require even more steps to complete. The theoretical horizon of policies with a discount factor of $\gamma=0.99$ is very large, exceeding well over 100 steps. It would be interesting to see how far one can push this limit in practice.

Another way to increase the task complexity would be to design a task and reward function, for which the agent has to work through penalties to get a high final reward. This task would test the capabilities of the training algorithm to escape local maxima of the reward function.

Other topics for future work include the investigation of the performance of the model for problems which require more action types. Reintroducing grasping and adding other motion primitives like placing grasped objects come to mind. While the UCB exploration did not improve performance in our case, we think that the technique would warrant further investigation in these scenarios.

Finally, it would of course be necessary to carry out future experiments on a real robot. Experiments in simulated environments serve as a good first step, but can't replace research in the real world. Two possibilities would be to transfer behaviour learnt in a simulation to a real world scenario or to perform training in the real world. Challenges that could come up include noisy camera images and generating enough experiences to train the agent.

\cleardoublepage


\backmatter

\iflanguage{english}{%
  \bibliographystyle{abbrvnat}%
}{%
  \bibliographystyle{dinat}%
}
\bibliography{bibliography}
\cleardoublepage

\appendix


\chapter{Appendix}

\begin{landscape}
\begin{table}[ht!]
\caption{Parameters and test results for the Hourglass models}
\label{app:results}
\centering
\begin{tabular}{c|cccc||cccccc}
\toprule
Name & $\gamma$ & $\gamma$-schedule & $\mathcal{L}$ & UCB $c$ &Complete \% & $\overline{G}_\textrm{max}$ & $\sigma(G_\textrm{max})$ & Change \% & $\overline{N}_A$ & $\sigma(N_A)$ \\
\midrule
& 0.8  & No  & Huber & 0 & 67.5\% & 5.175 & 1.48 & 78.7\% & 26.59 & 16.12 \\
$\textrm{M}_{\gamma=0.8}$ & 0.8  & No  & Huber & 1 & 90.0\% & 5.65  & 1.25 & 87.4\% & 29.00 & 16.10 \\
& 0.8  & No  & MSE   & 0 & 92.5\% & 5.825 & 0.71 & 90.5\% & 24.51 & 15.35 \\
& 0.8  & No  & MSE   & 1 & 80.0\% & 5.525 & 1.11 & 87.4\% & 23.44 & 12.49 \\
& 0.99 & No  & Huber & 0 & 85.0\% & 5.675 & 0.86 & 87.2\% & 29.21 & 14.64 \\
$\textrm{M}_\textrm{best}$ & 0.99 & No  & Huber & 1 & \textbf{95.0\%} & \textbf{5.975} & 0.16 & \textbf{95.5\%} & 22.89 & 9.86  \\
& 0.99 & No  & MSE   & 0 & 90.0\% & 5.9   & 0.30 & 90.6\% & 25.81 & 11.94 \\
& 0.99 & No  & MSE   & 1 & 87.5\% & 5.725 & 1.01 & 96.7\% & 23.91 & 9.12  \\
& 0.8  & Yes & Huber & 0 & 82.5\% & 5.65  & 0.92 & 90.5\% & 23.15 & 11.04 \\
& 0.8  & Yes & Huber & 1 & 82.5\% & 5.675 & 0.92 & 86.6\% & 22.52 & 9.74  \\
& 0.8  & Yes & MSE   & 0 & 92.5\% & 5.8   & 0.85 & 87.6\% & 26.24 & 17.00 \\
& 0.8  & Yes & MSE   & 1 & 92.5\% & 5.9   & 0.38 & 89.6\% & 22.95 & 12.49 \\
$\textrm{M}_{2^\textrm{nd}}$ & 0.99 & Yes & Huber & 0 & \textbf{95.0\%} & 5.95  & 0.22 & 92.6\% & 25.24 & 12.34 \\
$\textrm{M}_{\gamma-\textrm{sched}}$ & 0.99 & Yes & Huber & 1 & 90.0\% & 5.85  & 0.48 & 89.7\% & 30.47 & 17.10 \\
& 0.99 & Yes & MSE   & 0 & 92.5\% & 5.925 & 0.27 & 92.1\% & 22.73 & 12.43 \\
& 0.99 & Yes & MSE   & 1 & 92.5\% & 5.825 & 0.68 & 94.1\% & \textbf{21.97} & 11.52 \\
$\textrm{M}_{\gamma=0}$ & 0    & No  & Huber & 0 & 32.5\% & 4.6   & 1.52 & 81.8\% & 78.23 & 34.52 \\
& 0    & No  & Huber & 1 & 5.0\%  & 3.85  & 1.25 & 82.8\% & 38.50 & 0.71  \\
& 0    & No  & MSE   & 0 & 25.0\% & 4.275 & 1.66 & 86.3\% & 93.40 & 29.84 \\
& 0    & No  & MSE   & 1 & 17.5\% & 4.25  & 1.26 & 77.4\% & 42.00 & 23.41 \\
\bottomrule
\end{tabular}
\end{table}
\end{landscape}

\begin{figure}[!ht]
    \centering
    \subfigure{\includegraphics[width=0.4\textwidth, height=5.5cm]{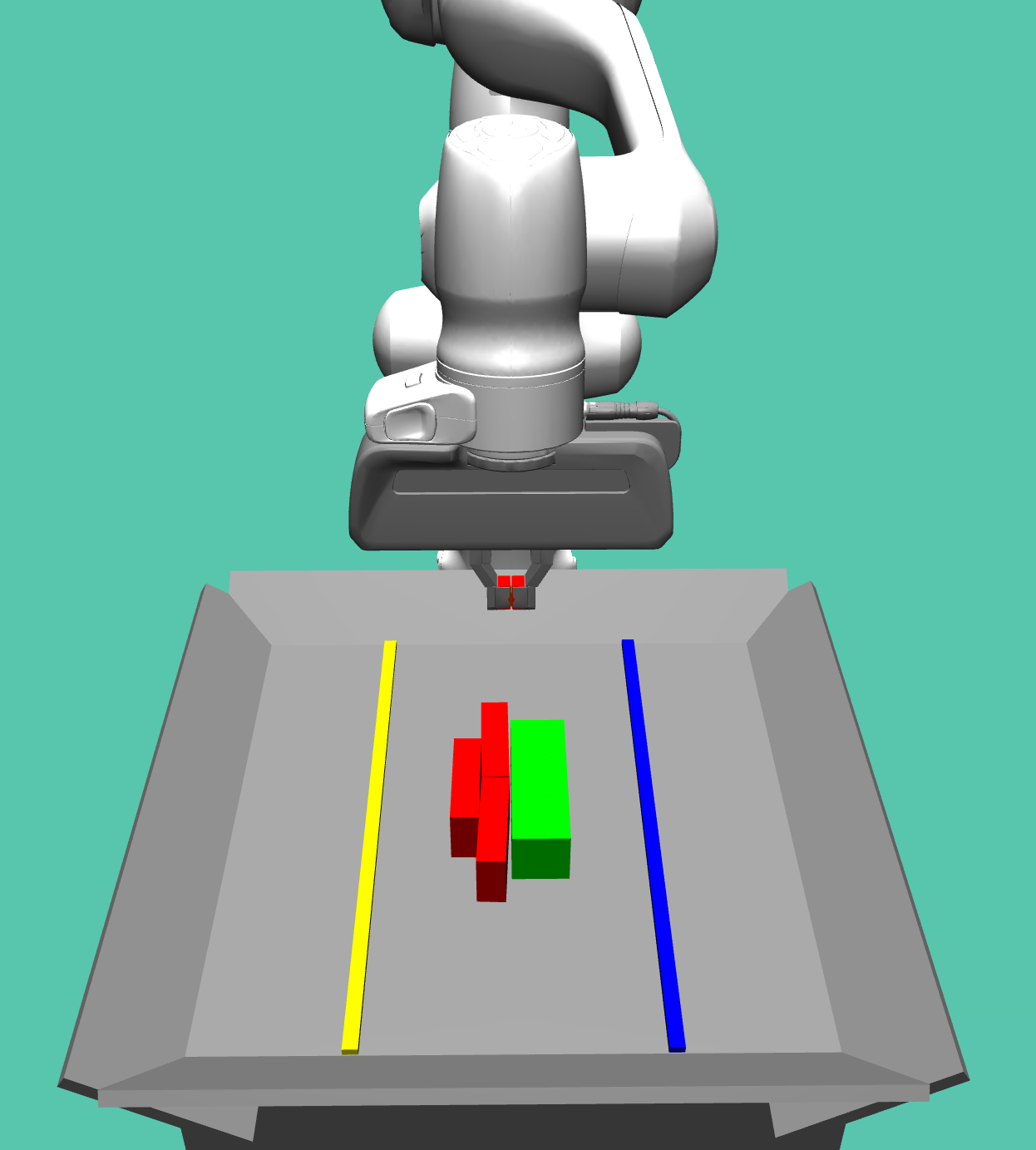}}
    \hspace{0.1\textwidth}
    \subfigure{\includegraphics[width=0.4\textwidth, height=5.5cm]{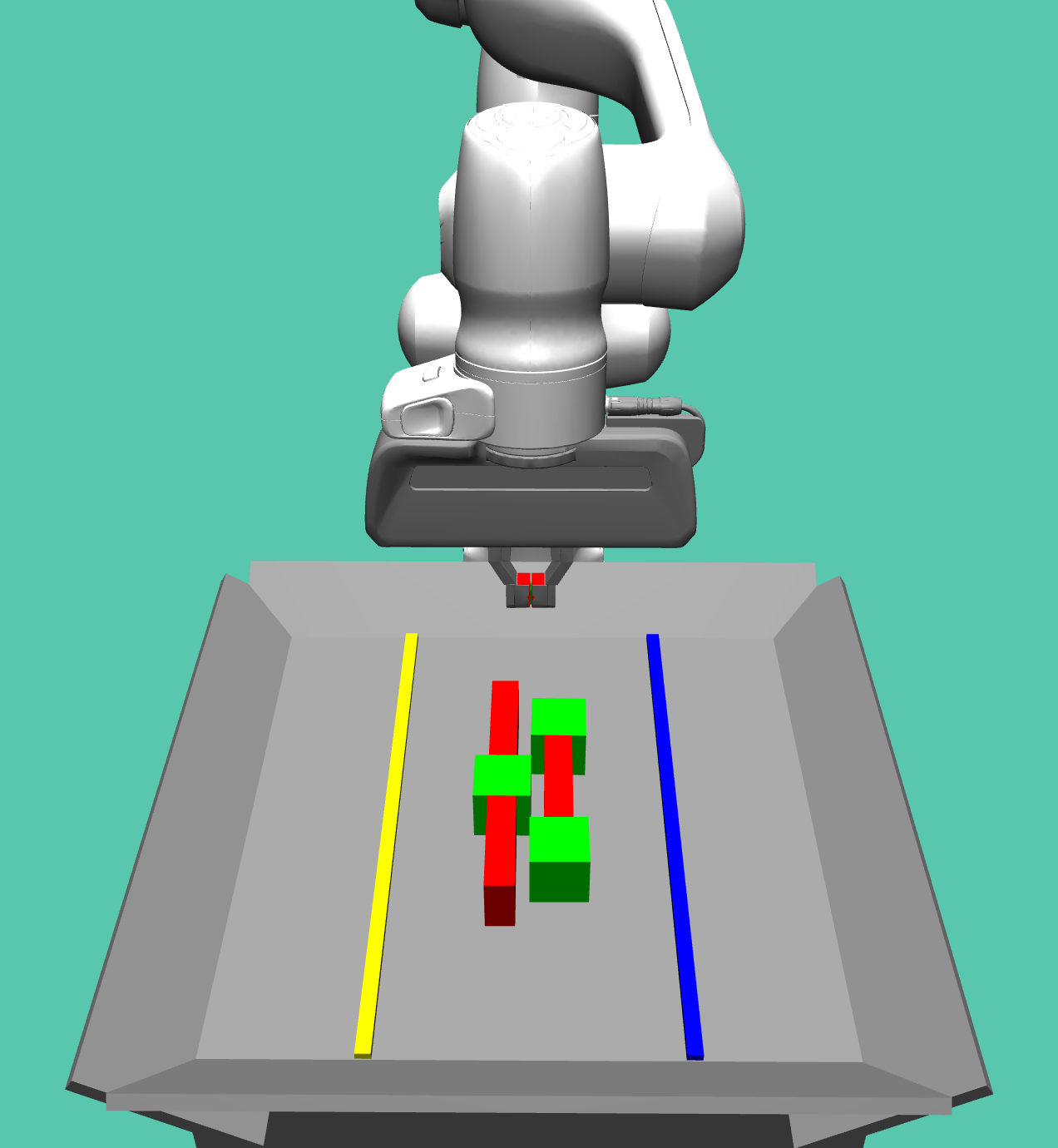}}
\end{figure}
\begin{figure}[!ht]
    \centering
    \subfigure{\includegraphics[width=0.4\textwidth, height=5.5cm]{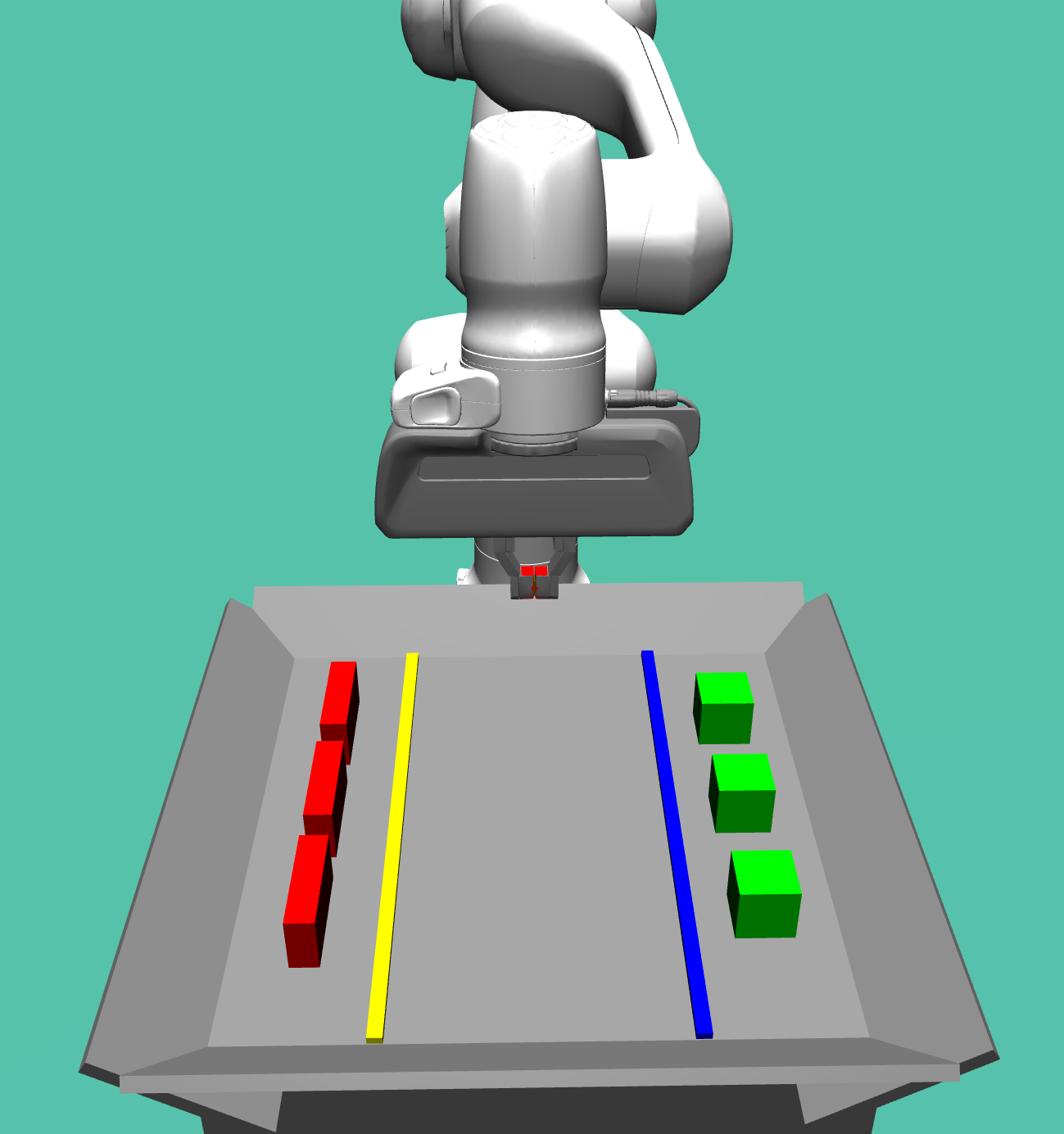}}
    \hspace{0.1\textwidth}
    \subfigure{\includegraphics[width=0.4\textwidth, height=5.5cm]{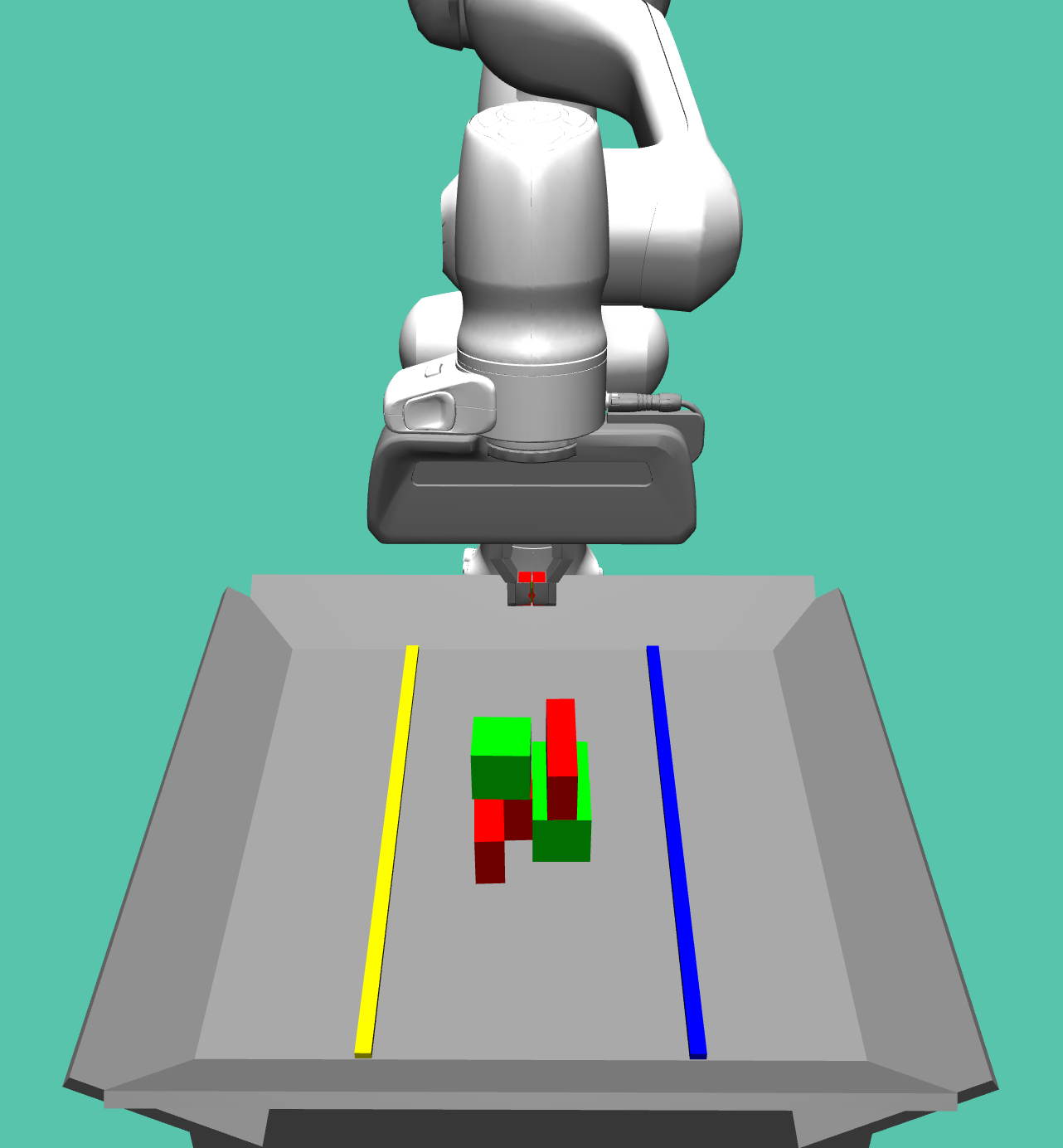}}
\end{figure}
\begin{figure}[!ht]
    \centering
    \subfigure{\includegraphics[width=0.4\textwidth, height=5.5cm]{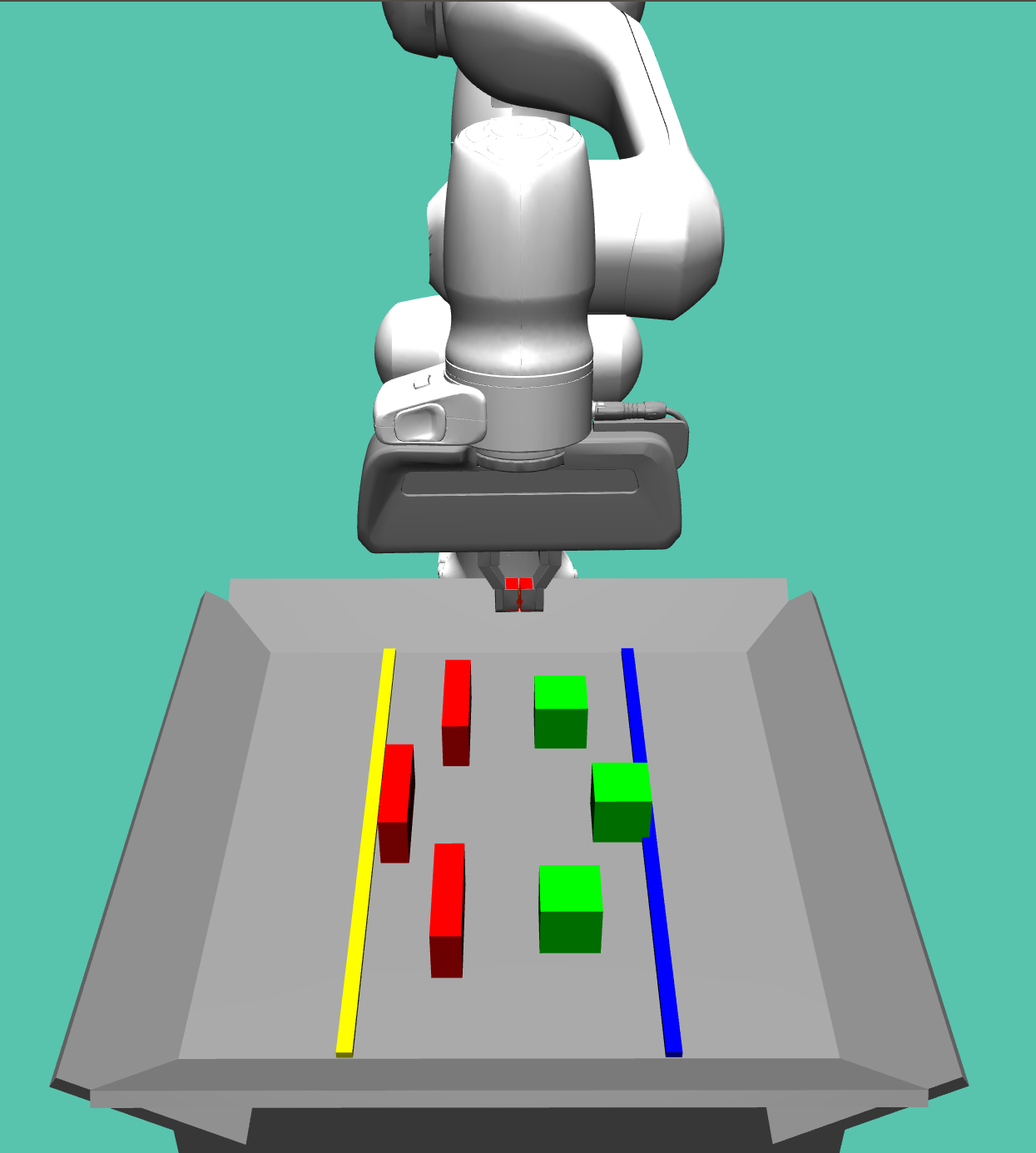}}
    \caption{Challenging arrangements.}
    \label{app:challenging} 
\end{figure}

\cleardoublepage

\end{document}